\documentclass{article}

\PassOptionsToPackage{numbers, compress}{natbib}

 \usepackage[preprint]{neurips_2026}
\makeatletter
\renewcommand{\@noticestring}{}
\makeatother

\usepackage[utf8]{inputenc} %
\usepackage[T1]{fontenc}    %
\usepackage{hyperref}       %
\usepackage{url}            %
\usepackage{booktabs}       %
\usepackage{amsfonts}       %
\usepackage{nicefrac}       %
\usepackage{microtype}      %
\usepackage{xcolor}         %
\usepackage{colortbl}       %
\definecolor{citeblue}{RGB}{31,78,165}
\hypersetup{colorlinks=true, citecolor=citeblue, urlcolor=citeblue, linkcolor=black}

\usepackage{amsmath}
\usepackage{amssymb}
\usepackage{mathtools}
\usepackage{amsthm}
\usepackage{xspace}
\usepackage{xurl}
\usepackage{graphicx}
\usepackage{wrapfig}
\usepackage{multirow}
\usepackage{makecell}
\usepackage{float}
\usepackage{longtable}
\usepackage{tabularx}
\usepackage{soul}
\usepackage{tablefootnote}
\usepackage{latexsym,placeins}
\usepackage{caption}
\usepackage{subcaption}
\usepackage{adjustbox}
\usepackage{changepage,threeparttable}
\usepackage{enumitem}
\usepackage{listings}
\usepackage[most]{tcolorbox}
\usepackage{algorithm}
\usepackage{algorithmic}
\usepackage[capitalize,noabbrev]{cleveref}
\usepackage{fontawesome5}
\usepackage{fancyhdr}

\definecolor{promptbg}{RGB}{245,245,255}
\newtcolorbox{promptbox}[1][]{
  colback=promptbg,
  colframe=black!20,
  fontupper=\small\ttfamily,
  boxrule=0.4pt,
  arc=2pt,
  left=6pt, right=6pt, top=4pt, bottom=4pt,
  #1
}
\newcommand{\best}[1]{\textbf{#1}}
\newcommand{\fc}{\cellcolor{blue!6}}
\definecolor{goldtint}{RGB}{255,243,205}
\definecolor{silvertint}{RGB}{230,230,230}
\definecolor{bronzetint}{RGB}{240,220,200}
\newcommand{\cgold}{\cellcolor{goldtint}}
\newcommand{\csilver}{\cellcolor{silvertint}}
\newcommand{\cbronze}{\cellcolor{bronzetint}}

\newcommand{\adobelogo}{\textsuperscript{\includegraphics[height=1.3em]{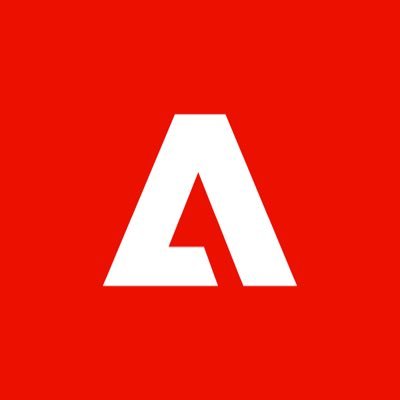}}}
\newcommand{\adobelogoheader}{\raisebox{-.1\height}{\includegraphics[height=1.3em]{logos/adobe-logo.jpg}}}
\newcommand{\iiitdlogo}{\textsuperscript{\includegraphics[height=1.0em]{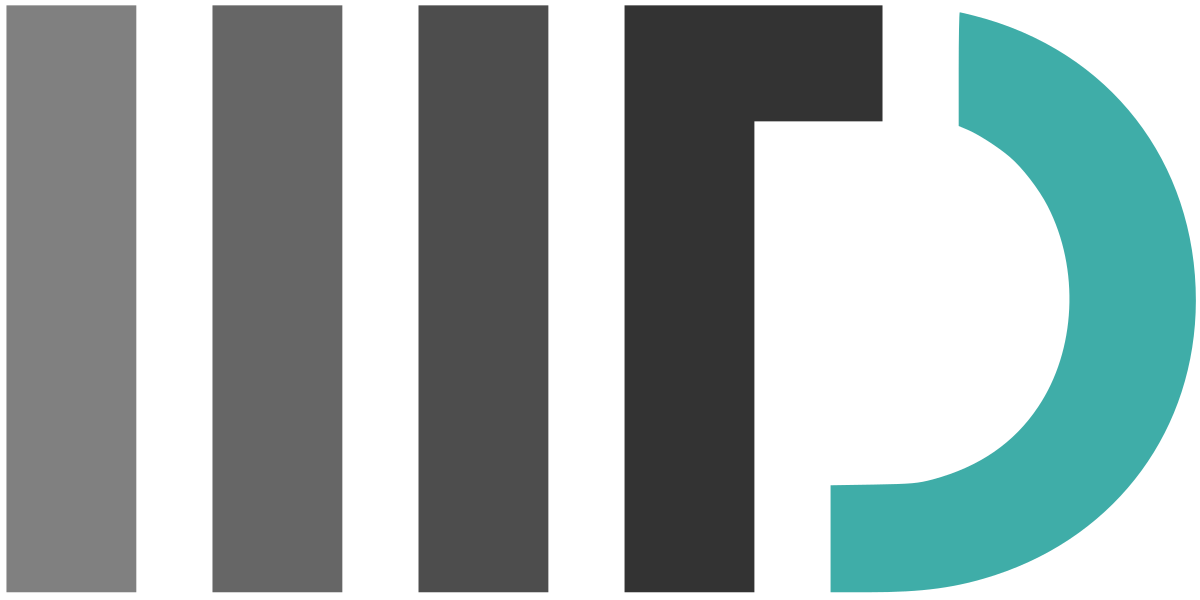}}}

\title{Bridging Expert Knowledge and Automated \\ Feature Engineering via Self-Evolution}

\vspace{-1em}

\author{%
\textbf{Varun Khurana}\adobelogo \hspace{0.8em} \textbf{Vijval Ekbote}\iiitdlogo \hspace{0.8em} \textbf{Vashu Chauhan}\iiitdlogo\\[2pt]
\textbf{Yaman K Singla}\adobelogo \hspace{0.8em} \textbf{Rajiv Ratn Shah}\iiitdlogo \hspace{0.8em} \textbf{Balaji Krishnamurthy}\adobelogo\\[3pt]
\adobelogo~Adobe Media and Data Science Research, \hspace{0.5em} \iiitdlogo~IIIT-Delhi\\[2pt]
\small{\faGlobe~\href{https://behavior-in-the-wild.github.io/fest.html}{\texttt{behavior-in-the-wild.github.io/fest}}}\\[1pt]
\small{\faEnvelope~\href{mailto:behavior-in-the-wild@googlegroups.com}{\texttt{behavior-in-the-wild@googlegroups.com}}}
\vspace*{-5pt}
}

\begin{document}

\pagestyle{fancy}
\fancyhf{}
\fancyhead[L]{\small\adobelogoheader~Adobe Media \& Data Science Research}
\fancyfoot[C]{\thepage}
\renewcommand{\headrulewidth}{0.4pt}
\renewcommand{\footrulewidth}{0pt}
\fancypagestyle{plain}{
  \fancyhf{}
  \fancyhead[L]{\small\adobelogoheader~Adobe Media \& Data Science Research}
  \fancyfoot[C]{\thepage}
  \renewcommand{\headrulewidth}{0.4pt}
  \renewcommand{\footrulewidth}{0pt}
}

\maketitle

\vspace{-1em}
\begin{abstract}
\vspace{-1em}

In high-stakes settings such as brand compliance, clinical care, and content moderation, machine learning cannot be deployed as opaque oracles: practitioners must inspect the features driving model decisions, and models must leverage the expert documentation governing these domains. In practice, the data arrives as unstructured content, and features extracted from it must be interpretable, discriminative, and aligned with what experts consider important. Existing methods fall short: they target tabular inputs, lack demonstrated expert alignment, and cannot operationalize qualitative criteria such as ``maintain professional tone'' into precise features. We present FEST (Feature Engineering with Self-evolving Trees), combining dual-stream feature generation (semantic and deterministic), semantic deduplication, and tree-guided iterative evolution to discover auditable features from raw text and images. FEST leads in 17 of 20 classifier-task combinations across brand classification, content authenticity detection, and stress detection, with a mean gain of 4.2 pp over the strongest baseline across five classifiers. An LLM-as-judge evaluation shows FEST achieves 60--80\% coverage of expert-designed brand features at strict semantic-alignment thresholds, corroborated by a human expert study rating features highly on relevance, clarity, and actionability. When seeded with expert guidelines, FEST refines qualitative criteria into operational features, improving accuracy by 6--12 pp on average across brands. To enable systematic evaluation of expert alignment in automated feature engineering, we release BrandGuide, the first dataset pairing expert-designed features with 1M+ assets across 2{,}683 brands. By grounding feature engineering in expert knowledge, FEST opens a practical pathway for interpretable ML in domains demanding human oversight.
\end{abstract}

\vspace{-1em}

\section{Introduction}
\label{sec:intro}

\vspace{-1em}

Practitioners cannot deploy machine learning systems they cannot interrogate. In high-stakes domains such as clinical decision support, content moderation, and brand compliance, machine learning and domain expertise must operate in conjunction: decisions affect people in ways that demand both automation at scale and expert oversight against the standards practitioners maintain \citep{rudin2019stop}. A misjudged advertising campaign can damage a brand and trigger public backlash and regulatory scrutiny \citep{cresci2016matchcom}; a clinical risk model whose decision criteria a physician cannot verify against medical knowledge cannot be trusted; a content classifier whose flagging criteria a moderator cannot audit cannot be deployed. This requirement runs in two directions. First, the features driving automated decisions must be ones experts can inspect, validate, and recognize as domain-relevant. Second, when experts have already documented domain criteria (brand style guides, clinical protocols, editorial standards), automated systems must ingest and apply this accumulated knowledge rather than discover features from scratch. Features are the natural interface for both directions: they are what models operate on, and what experts can match against their own specifications. \looseness=-1

This makes feature engineering the critical bridge between automated systems and domain expertise. The core technical challenge has two parts: producing features from raw unstructured data that are simultaneously discriminative, interpretable, and aligned with what experts regard as meaningful; and ingesting existing expert documentation when available. Post-hoc explanation methods (LIME, SHAP, GradCAM; \citealp{ribeiro2016should, lundberg2017unified, selvaraju2017grad}) are not a substitute: they produce approximate attributions over a black-box model's internal representations, not features that practitioners can compare against expert specifications. Producing such features automatically, with empirical evidence of expert alignment, is the problem this paper addresses. Even measuring this alignment requires ground-truth expert features paired with the data they describe, a benchmark that did not exist prior to this work (\S\ref{sec:dataset}). \looseness=-1

Existing automated feature engineering methods do not meet this requirement (Appendix~\ref{app:related_work}). Classical systems such as AutoFeat and OpenFE \citep{horn2019autofeat, zhang2023openfe} search predefined transformations over tabular columns and cannot operate on raw text or images. Recent LLM-based approaches narrow part of the gap: FeatLLM \citep{han2024large} and LLM-FE \citep{abhyankar2025llm} use language models for feature discovery but remain restricted to tabular inputs, while Felix \citep{malberg2024felix} extends to unstructured text using single-pass generation without iterative refinement. Crucially, all of these systems optimize for downstream accuracy alone: they do not demonstrate that the features they discover match what experts consider important, and they provide no mechanism for incorporating existing expert documentation. \looseness=-1

The second part of the challenge arises when expert documentation is available, which is common: brand managers maintain style guides, clinicians follow diagnostic protocols, content platforms publish editorial standards. These specifications encode valuable domain knowledge but are expressed as high-level qualitative criteria (``maintain professional tone'' or ``use high-quality 
product images'') rather than reproducible, measurable features. LLMs produce inconsistent outputs when directly prompted with such ambiguous criteria \citep{zheng2023judging}. Operationalizing expert knowledge requires transforming qualitative guidelines into precise features grounded in empirical data: a system must learn what ``professional tone'' concretely means by observing examples that satisfy or 
violate it. No prior system has demonstrated both expert-aligned feature discovery from unstructured data and the ability to operationalize expert documentation. \looseness=-1

We present \textbf{FEST} (Feature Engineering with Self-evolving Trees), an iterative framework that generates semantic features (LLM-assessed, e.g., ``professional tone'') and deterministic features (executable code, e.g., emoji count) from contrastive sample pairs, then consolidates and refines them based on discriminative power. This same engine supports two complementary modes. \textit{Without expert seeding}, FEST discovers features from unstructured data that achieve 60--80\% coverage of expert-authored specifications under strict LLM-as-judge evaluation, corroborated by a human expert study (above 3.8/5 on relevance, clarity, actionability). \textit{With expert guidelines}, FEST operationalizes ambiguous criteria into precise definitions while discovering complementary patterns, improving accuracy by 6--12 pp on average across brands. Across brand classification (text/images), content authenticity detection, and stress detection, FEST leads in 17 of 20 classifier-task combinations (mean gain 4.2 pp), with ablations confirming complementary contributions of the semantic and deterministic streams. Even measuring expert alignment requires ground-truth expert features paired with the data they describe. To spur research in this direction, we release \textbf{BrandGuide}, the first dataset pairing expert-designed features with unstructured content (\S\ref{sec:dataset}). \textbf{Our key contributions:}
\vspace{-2pt}
\begin{enumerate}[leftmargin=11pt, itemsep=0.5pt, topsep=2pt]
    \item \textbf{Problem formalization}: We formalize a deployment-critical problem: producing interpretable features from unstructured data that domain experts recognize as meaningful, and operationalizing expert documentation when available. We bring this problem to the community and propose expert alignment as a measurable objective for automated feature engineering. \looseness=-1
    \item \textbf{BrandGuide dataset}: To enable systematic evaluation of expert alignment in automated feature engineering, we release \textbf{BrandGuide}, the first dataset pairing expert-designed features with unstructured content: 1M+ assets across 2,683 brands, 80 sectors, and 103 regions.
    \item \textbf{FEST framework}: We propose FEST, combining dual-stream feature generation, semantic deduplication, and tree-guided evolution. FEST leads in 17 of 20 classifier-task combinations across five classifiers (mean gain 4.2 pp), while maintaining interpretability.
    \item \textbf{Expert alignment}: FEST achieves 60--80\% coverage of expert features under strict LLM-as-judge thresholds (Felix drops to 0\% on certain brands). A human expert study corroborates this (above 3.8/5 on relevance, clarity, actionability).
    \item \textbf{Expert operationalization}: With expert seeds, FEST operationalizes qualitative criteria into more precise features, improving accuracy by 6--12 pp on average across brands.
\end{enumerate}

\section{Problem Formulation}

Given training data $\mathcal{D}_{\mathrm{train}} = \{(x_i, y_i)\}_{i=1}^n$ with raw inputs $x \in \mathcal{X}$ and labels $y \in \mathcal{Y}$, we seek an optimal feature set $\boldsymbol{\phi}^\star = \{\phi_1, \ldots, \phi_{|\boldsymbol{\phi}|}\}$ from hypothesis space $\Phi$ of feature functions $\phi : \mathcal{X} \rightarrow \mathbb{R}$:
\begin{equation}
\begin{aligned}
\boldsymbol{\phi}^\star
\;=\;
\arg\max_{\boldsymbol{\phi} \subset \Phi} \quad
& \mathcal{J}\!\left(
f,\, \boldsymbol{\phi} \,;\, \mathcal{D}_{\mathrm{train}}
\right) \\
\text{s.t.} \quad
& \forall\, \phi_j \in \boldsymbol{\phi}:\ \phi_j \text{ is interpretable}
\end{aligned}
\end{equation}
where $f : \mathbb{R}^{|\boldsymbol{\phi}|} \rightarrow \mathcal{Y}$ is a downstream classifier operating on feature representation $\boldsymbol{\phi}(x) = [\phi_1(x), \ldots, \phi_{|\boldsymbol{\phi}|}(x)]$, and $\mathcal{J}$ measures empirical accuracy:
$
\mathcal{J}(f, \boldsymbol{\phi};\, \mathcal{D}_{\mathrm{train}})
\;=\;
\frac{1}{|\mathcal{D}_{\mathrm{train}}|}
\sum_{(x,y) \in \mathcal{D}_{\mathrm{train}}} \mathbb{I}\!\left(f(\boldsymbol{\phi}(x)) = y\right).
$
Notably, the interpretability constraint is what distinguishes this formulation from standard accuracy maximization: each $\phi_j$ must be expressible as a natural-language predicate or a short executable function, ensuring the resulting model is auditable by domain practitioners. Feature bank size is not imposed as a hard constraint but is empirically stabilized through semantic clustering (\S\ref{sec:dedup}).

\textbf{Expert seeding.} When expert-authored guidelines $G = \{g_1, \ldots, g_m\}$ are available, the hypothesis space $\Phi$ is seeded with features derived from $G$. The system should operationalize these qualitative specifications into measurable features and discover complementary patterns beyond the documentation. The optimization objective remains unchanged (\S\ref{sec:expert_knowledge}). \looseness=-1

\textbf{Distinction from classical AFE.} Classical systems \citep{horn2019autofeat, zhang2023openfe} operate on tabular inputs $\mathcal{X} = \mathbb{R}^m$ with a fixed transformation grammar over predefined columns. Here, $\mathcal{X}$ consists of raw unstructured inputs (text, images) with no column structure, and $\Phi$ is the open-ended space of interpretable predicates. Even \textit{proposing} plausible features requires a generative model, converting an intractable search into a guided discovery process. \looseness=-1

\textbf{Terminology.} \textit{Expert guidelines} are qualitative criteria authored by practitioners (e.g., ``use inclusive language''). \textit{Features} are operational characteristics: either natural-language descriptions assessed via LLM confidence or executable functions. \textit{Feature encodings} are the numerical scores from applying features to samples; \textit{feature representations} are the concatenated vectors fed to the classifier. \looseness=-1

\vspace{-10pt}
\section{Methodology}

FEST operates in two modes corresponding to the two parts of the problem formulated above (\S\ref{sec:intro}). In \textit{discovery mode}, the system receives only labeled data and discovers features from scratch. In \textit{expert-seeded mode}, the system additionally receives expert-authored guidelines that initialize the feature bank, which FEST then refines and augments through the same iterative process. The architecture is identical in both modes; the only difference is initialization. We describe each component below.

\begin{figure}[t]
  \centering
  \includegraphics[width=\linewidth]{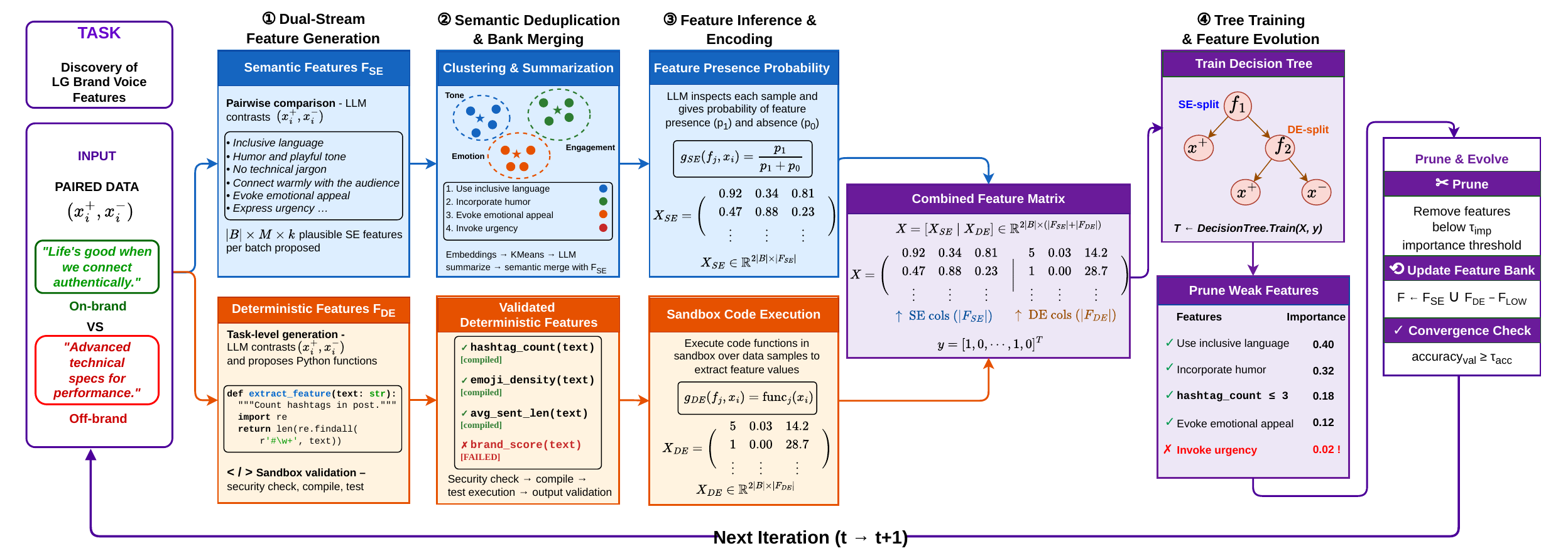}
  \caption{\small{One iteration of FEST on LG brand voice feature discovery. Given paired on-brand and off-brand samples, FEST executes two complementary streams: (i)~a \textit{semantic (SE)} stream, where an LLM generates pairwise discriminative hypotheses (e.g., ``inclusive language,'' ``emotional appeal''), followed by semantic clustering and deduplication; and (ii)~a \textit{deterministic (DE)} stream producing executable Python functions (e.g., \texttt{num\_hashtags}, \texttt{count\_punctuations}). Semantic features are encoded via LLM exponentiated log-probability confidence scores; deterministic features yield numeric values through execution. The concatenated representation trains a decision tree whose importance scores guide iterative pruning and refinement, yielding a compact, discriminative feature bank.}}
  \label{fig:arch_diagram}
  \vspace{-12pt}
\end{figure}

FEST iteratively discovers and refines $\boldsymbol{\phi}$ through LLM-guided generation and tree-based selection. We now describe FEST's methodology, organized around three key architectural innovations: (1) dual-stream discovery generating both semantic and deterministic features, (2) semantic consolidation via conditional embeddings and clustering, and (3) tree-guided iterative evolution using importance-based pruning. We explain each of them next. Figure~\ref{fig:arch_diagram} illustrates the complete pipeline, and Algorithm~\ref{algo:algo} provides the formal pseudocode. \looseness=-1

\subsection{Pairwise Comparison for Relative Discrimination}
Rather than analyzing samples in isolation, FEST discovers features through pairwise comparisons of positive and negative instances $\mathcal{P} = \{(x_i^+, x_i^-)\}$, which naturally filters common attributes and surfaces discriminative characteristics \citep{geng2025delta}. Pairwise comparison is used only for feature \textit{discovery}; feature \textit{encoding} (\S\ref{sec:feature_encoding}) scores each sample independently, ensuring generalization to unseen instances. See Appendix~\ref{app:pairwise_motivation} for extended motivation. \looseness=-1

\subsection{Iterative Feature Discovery and Refinement}
FEST processes training pairs in sequential batches, with each iteration comprising four stages: dual-stream feature generation, semantic deduplication, feature inference and encoding, and tree-based evolution. We initialize with feature bank $F = \emptyset$ (or with expert-designed features when available, see Section~\ref{sec:expert_knowledge}) and importance history $H_{importance} = \emptyset$, then iterate until convergence (validation accuracy exceeds threshold $\tau_{accuracy}$) or data exhaustion. \looseness=-1

\subsubsection{Dual-Stream Feature Generation}
Human-interpretable features span a spectrum from perceptual patterns requiring interpretation (``professional tone'', ``natural daylight'') to precisely measurable quantities (``image brightness'', ``capitalization ratio''). The key distinction is epistemological: some features can only be assessed through judgment, while others are deterministically computable by code. To capture both, FEST generates two complementary feature types in each iteration $t$:

\textbf{Semantic (SE) Features} capture perceptual, interpretive characteristics that require judgment to assess. For each pair $(x^+, x^-)$ in batch $B$, we prompt an LLM with multiple templates to generate $M$ features explaining what differentiates $x^+$ from $x^-$. This pairwise prompting occurs for every pair, creating a large pool $F_{SE}$ of candidate features that capture diverse perspectives on discriminative patterns.

\textbf{Deterministic (DE) Features} capture objective, precisely measurable characteristics through executable Python functions. Given the overhead of code generation and validation, we operate at task-level rather than pair-level: the LLM analyzes a small subset of pairs from $B$ (typically 5) along with the task description to propose general deterministic features (e.g., ``image aspect ratio'', ``emoji count''). Each proposed feature is implemented as Python code, executed in a sandbox on validation samples, and iteratively refined if errors occur. Task-level generation produces focused set $F_{DE}$ of general features rather than pair-specific measurements, and the LLM's awareness of previously generated deterministic features prevents redundancy across iterations.

This dual-stream architecture provides complementary representations: semantic features capture nuanced perceptual patterns (e.g., ``emotional storytelling,'' ``natural daylight'') while deterministic features provide precise quantitative measurements (e.g., capitalization ratio, color saturation), together enabling richer feature spaces than either alone. At all LLM prompting stages (generation, summarization, encoding), we explicitly instruct the model to exclude superficial brand identifiers (names, logos, hashtags) to prevent shortcut learning (see Appendix~\ref{app:prompts} for exact constraints). \looseness=-1

\subsubsection{Semantic Deduplication and Bank Merging}
\label{sec:dedup}
The semantic stream can generate thousands of features in $F_{SE}$ across pairs in a batch, many expressing identical concepts through varied phrasing ("uses informal language", "casual tone", "conversational style"). Without deduplication, this inflates dimensionality and dilutes signal. We perform semantic consolidation through three steps: (1) compute conditional embeddings of all features in $F_{SE}$ conditioned on the task description, by prepending the task description as an instruction prefix before encoding with Qwen3-Embedding-4B~\citep{zhang2025qwen3}, making the embedding space domain-aware so that features with similar semantic roles cluster together regardless of surface phrasing, (2) cluster semantically similar features using K-means, and (3) prompt an LLM to summarize each cluster into a single representative feature preserving the core concept. This reduces redundancy while amplifying true signal by unifying equivalent features, producing deduplicated set $\bar{F}_{SE}$. \looseness=-1

These deduplicated semantic features $\bar{F}_{SE}$ are then merged with existing semantic features in feature bank $F$ through semantic similarity checking to prevent duplication with previous iterations. Deterministic features $F_{DE}$ require no explicit deduplication: their task-level generation, small number per iteration and LLM context-awareness about prior features naturally prevent redundancy. This produces the current iteration's complete feature set: $F^{(t)} = F \cup \bar{F}_{SE} \cup F_{DE}$, where $|F^{(t)}| = |F_{SE}| + |F_{DE}|$ with $|F_{SE}|$ denoting total semantic features and $|F_{DE}|$ denoting total deterministic features. \looseness=-1

\subsubsection{Feature Inference and Encoding}
\label{sec:feature_encoding}
While pairwise comparisons guide feature \textit{discovery}, encoding scores each sample independently to ensure generalization. Each pair $(x_i^+, x_i^-)$ is split into individual samples. Semantic features are encoded via LLM log-probability confidence: the LLM evaluates whether feature $f_k$ is present or absent in sample $x$, and we compute a normalized confidence score from the output probabilities, yielding a continuous value in $[0,1]$ that captures LLM certainty (Appendix~\ref{app:encoding_math}). Deterministic features are encoded by executing the associated Python function directly. The two feature matrices are concatenated into a combined representation $\mathbf{X} = [\mathbf{X}_{SE} \mid \mathbf{X}_{DE}]$ for decision tree training. Decision trees handle the heterogeneous types (probability scores and numeric values) naturally via Gini impurity reduction without normalization. \looseness=-1

\subsubsection{Tree-Based Evolution and Convergence}
We train a decision tree classifier $\mathcal{T}$ on $(\mathbf{X}, \mathbf{y})$ and evaluate on a held-out validation set $\mathcal{P}_{val}$. Decision trees serve multiple critical purposes for FEST:

\textbf{Why Decision Trees?} Trees provide (1) transparent decision paths for practitioner inspection, (2) feature importance via impurity reduction that directly guides evolution, and (3) automatic threshold learning (e.g., ``emoji count $< 5$'') that operationalizes vague criteria into precise splits. \looseness=-1

\textbf{Evolution Mechanism}: After training $\mathcal{T}$, we extract feature importance vector $\mathbf{I}$ and update history $H_{importance}$ tracking scores across iterations. Features in $F^{(t)}$ with consistently low importance over the last three iterations (mean importance below threshold $\tau_{importance}$) are pruned, producing $F^{(t)'}$ which becomes the global feature bank for the next iteration: $F \leftarrow F^{(t)'}$. This history-based pruning prevents spurious removal from single-iteration noise, ensuring only genuinely weak features are discarded. The feature bank thus evolves: high-value features persist while weak features are pruned, focusing capacity on predictive patterns. \looseness=-1

\textbf{Convergence}: The algorithm terminates when validation accuracy exceeds $\tau_{accuracy}$ or all training pairs are processed. A held-out test set is used only for final evaluation. See Algorithm~\ref{algo:algo} for full procedural details. \looseness=-1

\subsection{Leveraging Expert Knowledge}
\label{sec:expert_knowledge}
\looseness=-1
When expert-designed features exist (e.g., brand style guidelines, clinical criteria), FEST initializes the feature bank $F$ with these features rather than starting from $\emptyset$. This enables three capabilities: (1) \textit{Validation}: the iterative process retains high-importance expert features while pruning weak ones. (2) \textit{Refinement}: data-driven hypotheses often produce more specific reformulations of expert principles; the decision tree selects whichever version is more discriminative (e.g., ``high-quality images'' $\rightarrow$ ``close-up shots with resolution $> 1000$ pixels showing product texture''). (3) \textit{Augmentation}: FEST discovers features beyond expert knowledge, surfacing patterns experts may not have articulated. \looseness=-1

\vspace{-5pt}
\section{BrandGuide Dataset}
\label{sec:dataset}

Evaluating expert alignment requires ground-truth expert features paired with the content they describe, yet no such benchmark exists. In practice, organizations employ brand strategists who formulate comprehensive guidelines codifying how brand assets must be constructed: color palettes with exact codes, typography hierarchies, logo placement rules, and tone-of-voice specifications. These represent deliberate, expert-validated design decisions, precisely the kind of domain knowledge that automated systems should be evaluated against. \looseness=-1

We release \textbf{BrandGuide}, pairing these expert specifications with corresponding brand assets. Our pipeline extracts structured guidelines across \textbf{2,683} brands from the web and retrieves corresponding imagery, connecting expert-defined rules to their practical instantiations. The dataset comprises \textbf{1M+} brand images and text across 80 sectors, 103 regions, and 28 languages (2014--2025). See Appendix~\ref{app:dataset} for collection methodology, examples, and statistics. \looseness=-1

\vspace{-2ex}
\section{Experiments}
\vspace{-1.5ex}
\subsection{Tasks and Datasets}
\vspace{-1ex}
We evaluate FEST across multiple tasks demonstrating its effectiveness in high-stakes domains requiring interpretable features:

\begin{enumerate}[left=0pt, labelsep=0.5em]
    \vspace{-1ex}
    \item \textbf{Brand Classification}:
    Brand consistency is critical for business success: consistent brand presentation increases revenue by 10--20\% on average \citep{lucidpress2019brand}, while strong brand identity drives customer recognition, loyalty, and competitive differentiation \citep{acar2024role, keller2003brand}. However, maintaining consistent brand identity across marketing campaigns is challenging as inconsistent messaging damages brand equity and customer trust. We classify social media content as on-brand or off-brand for 5 brands (Porsche, Adobe, Emirates, Louis Vuitton, Pizza Hut) spanning automotive, technology, aviation, luxury fashion, and food service sectors, using posts from the EngagingImageNet dataset \citep{khurana2025measuring}. Both modalities are evaluated: \textit{text} (captions, based on linguistic style and tone) and \textit{images} (promotional visuals, based on aesthetics and composition). Posts from competing brands in the same sector serve as off-brand samples. \looseness=-1
    \vspace{-1ex}
    \item \textbf{Content Authenticity Detection}:
    Distinguishing AI-generated from human-written content is increasingly important for content moderation and information integrity. We evaluate FEST on detecting whether a story is written by humans or AI systems using the GPT-GC dataset from \citet{zhou2024hypothesis}. \looseness=-1
    \vspace{-1ex}
    \item \textbf{Stress Detection}:
    Identifying psychological stress from text enables mental health support applications. We evaluate FEST on detecting whether Reddit post authors exhibit stress using the Dreaddit dataset \citep{turcan2019dreaddit}, which contains posts from stress-related and neutral subreddits.

\end{enumerate}

\vspace{-2ex}
\textbf{Data Splitting.} For each task, data is pre-partitioned into three disjoint sets: (1) training pairs for iterative feature generation, (2) a validation set $\mathcal{P}_{val}$ used inside the FEST loop for convergence monitoring, and (3) a held-out test set $\mathcal{P}_{test}$ used only for final evaluation and reported in all results tables. \looseness=-1

\subsection{Baseline Methods}
We benchmark FEST's performance against baseline methods. There are two main components in each method; the feature generator and the classifier. \\
\textbf{Feature Generators: }For feature generation, we employ 3 different backbones. The first is an LLM that discovers features given just the task description. The second backbone is similar, with the only difference being that the LLM is also passed few-shot examples. The third backbone is Felix \citep{malberg2024felix}, which generates features from pairs, clusters them, and creates feature value vectors using an LLM. For the zero-shot and few-shot cases, feature vectors are obtained using the same feature inference pipeline as FEST. \looseness=-1 \\
\textbf{Downstream Classifiers: }We use the feature vectors obtained from each of the above backbones to train different downstream classifiers, namely, decision tree (DT), logistic regression (LR), random forest (RF), multi-layer perceptron (MLP), and XGBoost (XGB). Classifier hyperparameters are fixed identically across all feature generators so that accuracy differences isolate feature quality rather than confounding it with classifier optimization. Using multiple classifiers demonstrates that FEST's feature quality generalizes beyond tree-based learners.

All methods use GPT-4o-mini \citep{openai2024gpt4omini} as the LLM backbone (temperature settings in Appendix~\ref{app:hyperparameters}). Since all baselines use the identical LLM on identical content, any data contamination benefits all methods equally; FEST's consistent gains across both brand and non-brand datasets confirm the improvements are methodological (Appendix~\ref{app:contamination}). \looseness=-1

\section{Results and Discussion}
\vspace{-0.5em}

We evaluate both parts of the problem (\S2): whether FEST discovers discriminative, expert-aligned features, and whether it can operationalize expert documentation. We organize the evaluation into three tiers: \looseness=-1

\vspace{-0.5em}

\begin{enumerate}[leftmargin=16pt, noitemsep, topsep=2pt]
\item \textbf{Task performance} (\S\ref{sec:classification_results}): Classification accuracy across four tasks, five classifiers, and ablations isolating the contribution of each feature stream. This establishes that FEST features are discriminative.
\item \textbf{Expert alignment} (\S\ref{sec:expert_validation}): For brand classification, where expert guidelines exist, we measure whether FEST's discovered features align with expert-authored specifications through (a) automated coverage via an LLM-as-judge protocol and (b) a human expert study rating feature relevance, clarity, and actionability.
\item \textbf{Expert knowledge operationalization} (\S\ref{sec:expert_refinement}): A controlled experiment measuring FEST's ability to ingest, refine, and augment existing expert guidelines into more precise and discriminative feature sets.
\end{enumerate} \looseness=-1

\vspace{-1em}
\subsection{Task Performance}
\label{sec:classification_results}
\vspace{-1em}

\begin{wraptable}{r}{0.58\textwidth}
    \vspace{-14pt}
    \centering
    \caption{\small{Classification accuracy (\%) across tasks and downstream classifiers. FEST leads in 17 of 20 classifier-task combinations across brand classification (text and images), AI content detection, and stress detection. Results shown for Decision Tree (DT), Logistic Regression (LR), Random Forest (RF), MLP, and XGBoost (XGB) classifiers, all using GPT-4o-mini \citep{openai2024gpt4omini} as the LLM backbone. Bold indicates best performance per classifier-task combination. Brand classification results averaged over 5 brands; brand-wise breakdowns in Appendix~\ref{appendix:additional_results}.}}
    \label{tab:classification_results}
    \vspace{-4pt}
    \resizebox{\linewidth}{!}{%
    \begin{tabular}{cccccc}
    \toprule
    \multirow{2}{*}{\textbf{Clf.}} &
    \multirow{2}{*}{\textbf{Feature Generator}} &
    \multicolumn{4}{c}{\textbf{Task}} \\ \cline{3-6}
    & & \makecell{Brand Cl. \\ (Text)} & \makecell{Brand Cl. \\ (Images)} & \makecell{Cont\\ Auth.} & \makecell{Stress \\ Det.} \\ \toprule
    
    \multirow{4}{*}{DT}
     & Zero-Shot LLM     & 73.29  & 65.21 & 70.40 &  74.40\\
     & Few-Shot LLM     &  72.98 & 71.29  & 74.00 & 76.80 \\
     & Felix & 75.86  & 67.82  & 81.60 & 75.60 \\
     & \fc FEST (Ours) & \fc\textbf{81.70} & \fc\textbf{78.32} & \fc\textbf{91.20} & \fc\textbf{78.00} \\ \hline
    
    \multirow{4}{*}{LR}
     & Zero-Shot LLM     &  69.50  & 69.82    & 83.60 & 72.80 \\
     & Few-Shot LLM     &   72.96  &  73.05   & 60.80 &  67.20\\
     & Felix &  75.70   &  69.72   & \textbf{88.40} & 76.40 \\
     & \fc FEST (Ours) & \fc\textbf{81.59} & \fc\textbf{78.92} & \fc 84.00 & \fc\textbf{81.60} \\ \hline
    
    \multirow{4}{*}{RF}
     & Zero-Shot LLM     &  80.27 &  74.46   & 81.20 & 78.40 \\
     & Few-Shot LLM     & 83.17 &  78.71  & 86.80 &  77.60\\
     & Felix & 80.03 & 70.58    & 88.80 & \textbf{84.40} \\
     & \fc FEST (Ours) & \fc\textbf{85.11} & \fc\textbf{81.55} & \fc\textbf{97.20} & \fc 83.60 \\ \hline
    
    \multirow{4}{*}{MLP}
     & Zero-Shot LLM     & 75.96  & 70.28  & 78.40 & 65.60 \\
     & Few-Shot LLM     & 76.75  & 73.49  & 64.40 & 65.20 \\
     & Felix & 78.26  & 68.82  & 87.60 & \textbf{79.60} \\
     & \fc FEST (Ours) & \fc\textbf{81.57} & \fc\textbf{76.56} & \fc\textbf{88.40} & \fc 78.40 \\ \hline
    
    \multirow{4}{*}{XGB}
     & Zero-Shot LLM     & 79.11  & 73.02  & 85.20 & 75.20 \\
     & Few-Shot LLM     & 83.04  & 76.86  & 83.60 & 77.20 \\
     & Felix & 80.68  & 71.67  & 91.20 & 79.60 \\
     & \fc FEST (Ours) & \fc\textbf{84.58} & \fc\textbf{81.29} & \fc\textbf{94.40} & \fc\textbf{80.80} \\ \bottomrule
    \end{tabular}
    }
    \vspace{-10pt}
\end{wraptable}

Table~\ref{tab:classification_results} presents classification accuracy across four tasks using five classifiers and four feature generation methods. Brand classification results are averaged across 5 brands; brand-wise breakdowns appear in Appendix~\ref{appendix:additional_results} (Table \ref{tab:brandwise_classification_results_text}, \ref{tab:brandwise_classification_results_image}).

\textbf{FEST leads in 17 of 20 classifier-task combinations} (mean gain of 4.2 pp over the respective strongest baseline), with ablations confirming complementary contributions of both feature streams (Table~\ref{tab:classification_results}). Gains are consistent across brand classification on both modalities and content authenticity detection, where FEST's largest single margin reaches 8.4 pp over Felix (RF). While FEST wins on average, brand-wise analysis reveals outliers where baselines occasionally excel for specific brands (Appendix~\ref{appendix:additional_results}). On stress detection, Felix shows competitive performance, indicating semantic-only features may suffice for some psychological tasks. \looseness=-1

\textbf{Failure analysis.} FEST underperforms in 3 of 20 classifier-task combinations, all involving tasks where semantic-only features are sufficient. On stress detection, Felix leads with RF (84.4\% vs.\ 83.6\%) and MLP (79.6\% vs.\ 78.4\%), and on content authenticity with LR (88.4\% vs.\ 84.0\%). The common pattern: these are short-text tasks where deterministic features (sentence length, punctuation counts) add limited signal beyond semantic descriptions, and Felix's larger unconstrained feature pool captures marginal semantic variations. FEST's controlled feature space (30 features per iteration via K-means) trades exhaustive coverage for precision, a net positive on brand tasks but occasionally limiting on simpler domains. \looseness=-1

\textbf{Generalization across classifiers.} Despite using decision trees internally for evolution, FEST features transfer effectively to all five downstream classifiers (DT, LR, RF, MLP, XGBoost), demonstrating genuine discriminative patterns rather than tree-specific artifacts.

\textbf{Dual-stream features outperform semantic-only approaches.} Felix \citep{malberg2024felix}, the strongest baseline for unstructured data, generates only semantic features through single-shot LLM prompting. An ablation study (SE-only, DE-only, SE+DE) across 7 tasks confirms complementarity: SE+DE wins 11 of 14 DT+RF combinations. DE-only is always weakest, confirming that semantic features form the core signal while deterministic features add complementary measurable precision. Full ablation in Appendix~\ref{app:ablation_sede}. \looseness=-1

\textbf{Sources of gain.} FEST's improvements arise from the interplay of pairwise contrastive discovery, dual-stream generation, iterative importance-based pruning, and semantic deduplication (300+ candidates consolidated to 30 per iteration). \looseness=-1

\textbf{Robustness to shortcuts.} To verify FEST learns genuine voice/style features rather than exploiting brand identifiers, we generated synthetic off-brand content using GPT-4o-mini that matches each brand's topics but uses generic language. FEST achieves 84.4\% (Adobe), 79.8\% (LG), 91.2\% (Porsche) on this controlled setup, and the top features are purely stylistic (e.g., ``instructional tone,'' ``exclamation usage,'' ``sentence length variance''). See Appendix~\ref{app:synthetic} for details. \looseness=-1

\textbf{Stability and efficiency.} Across 3 independent seeds (5 brands $\times$ 5 classifiers), FEST produces consistent feature banks with standard deviations mostly below 3pp for text classification; image and smaller-dataset tasks show slightly higher variance driven by classifier sensitivity rather than feature instability (Appendix~\ref{app:variance}). K-means deduplication ($k$=30) is the stabilizing mechanism: stochastic LLM generation converges to consistent representative features after clustering and summarization. FEST averages \$0.10 per run vs.\ Felix's \$8.62 (86$\times$ cheaper) and completes in 15.9 minutes vs.\ 69.4 minutes (4.4$\times$ faster), consuming 91$\times$ fewer tokens. This efficiency stems from iterative pruning: features eliminated by decision tree importance are not re-encoded in subsequent iterations (Appendix~\ref{app:runtime}). \looseness=-1

\vspace{-1em}
\subsection{Expert Validation}
\label{sec:expert_validation}
\vspace{-1em}

\begin{wrapfigure}{r}{0.55\textwidth}
\vspace{-12pt}
\centering
\includegraphics[width=0.53\textwidth]{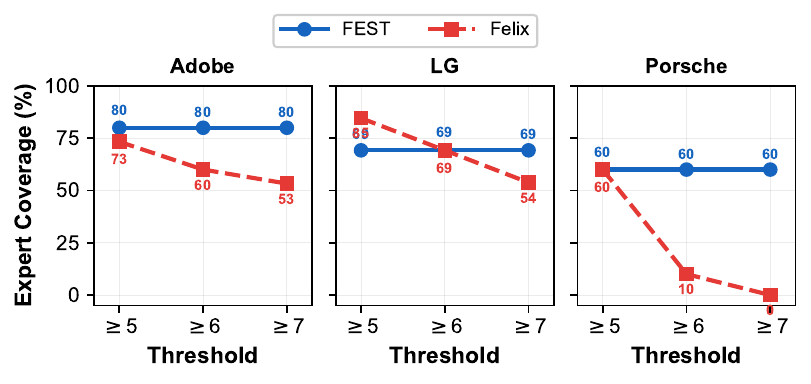}
\vspace{-6pt}
\caption{\small{LLM-as-judge expert coverage (\%) at thresholds 5--7 for three brands (Adobe, LG, Porsche). FEST (solid) maintains stable coverage across all thresholds, confirming strong semantic matches (7--9/10). Felix (dashed) appears comparable at threshold 5 but collapses at stricter thresholds, reaching 0\% for Porsche at $\geq$7. Full per-brand sweep in Appendix~\ref{app:llm_judge}.}}
\label{fig:coverage_plot}
\vspace{-10pt}
\end{wrapfigure}

Classification accuracy alone is insufficient for high-stakes deployment: discovered features must also align with what domain experts consider important. Expert brand guidelines are the authoritative specifications that practitioners use to ensure brand consistency, and commercial compliance platforms \citep{adobe2024genstudio,canva2024brandkit} ingest them as ground truth. When FEST-discovered features align with these guidelines, it provides direct evidence that FEST operates on the same semantic dimensions as human experts. We validate FEST's alignment through two complementary evaluations: automated coverage via an LLM-as-judge protocol, and a human expert study. \looseness=-1

\textbf{LLM-as-Judge Protocol.} For three brands (Adobe, LG, Porsche), we compare the top-20 features discovered by FEST and Felix against expert-designed brand voice guidelines. GPT-4o rates each (guideline, feature) pair on a 0--10 semantic alignment scale; a guideline is ``covered'' if any feature scores at or above the threshold. This provides interpretable coverage scores without reliance on embedding similarity cutoffs. \looseness=-1

\textbf{Coverage results.} At threshold $\geq 7$, FEST achieves 60--80\% coverage across brands and remains perfectly stable from threshold 5 through 7: the features covering a guideline score 7--9/10, not borderline, confirming coverage is not an artifact of threshold choice. Felix appears comparable at threshold 5, but collapses at higher thresholds, reaching 0\% on one brand at threshold 7 (Figure~\ref{fig:coverage_plot}). Full per-brand sensitivity and average alignment scores in Appendix~\ref{app:llm_judge}. \looseness=-1

\textbf{Why FEST achieves higher coverage.} Both methods use the same backbone LLM (GPT-4o-mini) and contrastive samples, so the coverage gap stems from what happens \textit{after} feature proposals: cluster summarization (vs.\ centroid picking), iterative pruning of generic features, and multi-stage language refinement each contribute to convergence toward expert-aligned formulations. See Appendix~\ref{app:coverage_analysis} for the full analysis. \looseness=-1

\textbf{Human expert study.} Domain experts rated 15 FEST-discovered features per brand on Relevance, Clarity, and Actionability (1--5 scale). Across two brands, all three dimensions score above 3.8/5, well above the midpoint. The expert reviewed both refined and newly discovered features as a single blinded pool, validating FEST's complete output. See Appendix~\ref{app:human_study} for the full protocol. The convergence of task performance, automated coverage, and practitioner ratings provides stronger evidence than any single metric alone. \looseness=-1

\vspace{-0.75em}
\subsection{Expert Knowledge Operationalization}
\label{sec:expert_refinement}
\vspace{-0.5ex}

Having established that FEST's discovery mode produces features aligned with expert knowledge (\S\ref{sec:expert_validation}), we evaluate the complementary capability: can FEST operationalize qualitative specifications into precise, measurable features and discover complementary patterns? We conduct a controlled experiment using brand style guidelines as seed features. \looseness=-1

\begin{wrapfigure}{r}{0.45\columnwidth}
\vspace{-1em}
\centering
\includegraphics[width=0.46\columnwidth]{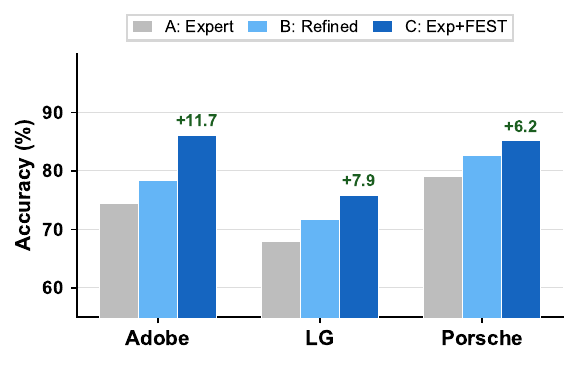}
\vspace{-1em}
\caption{\small{Expert feature refinement accuracy (averaged across DT, LR, RF, LLM classifiers) for text brand classification. A: expert features as-is, B: FEST features filtered to those aligned with experts (LLM-judge $\geq 7$), C: full Expert+FEST. The staircase pattern (A$\rightarrow$B$\rightarrow$C) confirms that refinement and augmentation each provide independent gains of 6--12 pp. Per-classifier breakdown in Table~\ref{tab:expert_refinement} (Appendix).}}
\label{fig:refinement_chart}
\vspace{-1em}
\end{wrapfigure}

\textbf{Experimental setup.} We extract expert-designed features from official brand style guides for three brands (LG, Porsche, Adobe), containing qualitative guidelines for brand identity (e.g., ``use high-quality images,'' ``maintain professional tone''). We evaluate FEST in two configurations: (1) \textit{Evaluation Mode}, where expert features are used directly without modification; and (2) \textit{Learning Mode}, where expert features initialize FEST's feature bank for full iterative discovery. Both configurations use identical train-test splits and evaluation protocols, isolating the effect of FEST's refinement mechanism. \looseness=-1

\textbf{Results.} Figure~\ref{fig:refinement_chart} summarizes the average-across-classifiers results (full per-classifier breakdown in Table~\ref{tab:expert_refinement}, Appendix). Three conditions disentangle the sources of improvement: (A) expert features alone, (B) expert refined, where FEST features are filtered to those aligned with expert features (LLM-judge score $\geq 7$), isolating refinement without augmentation, and (C) Expert+FEST, the full augmented output. FEST's learning mode (C) consistently outperforms expert features alone (A), achieving improvements of 6--12 pp on average across brands (DT, LR, RF). Critically, B$>$A in most cases: refinement alone improves over static expert features (e.g., Adobe DT: A=78.4\%, B=87.2\%), demonstrating that FEST operationalizes ambiguous criteria into more discriminative formulations. C$>$B in all cases: augmentation provides further independent gains (e.g., Porsche LR: B=86.4\%, C=90.0\%), confirming FEST discovers complementary features beyond what refinement captures. \looseness=-1

\textbf{Analysis of feature evolution.} FEST's feature bank evolves from expert seeds: some features are retained verbatim, ambiguous guidelines are refined into precise operational definitions, and novel features are discovered beyond the original documentation. See Appendix~\ref{app:feature_evolution} for detailed examples and a visualization of each category.

\vspace{-1em}

\section{Conclusion}
\vspace{-1em}
Automated feature engineering and domain expertise have largely operated in isolation: automated methods optimize for accuracy without demonstrating alignment with what experts consider important, while expert knowledge remains encoded in qualitative documentation that automated systems cannot ingest. FEST bridges this gap through two complementary modes, achieving 60--80\% coverage of expert specifications in discovery mode (corroborated by a human expert study) and 6--12 pp accuracy gains through expert operationalization, while leading in 17 of 20 classifier-task combinations.

More broadly, FEST shows that LLMs contain substantial latent domain knowledge that can be systematically extracted through the right architectural scaffolding: contrastive grounding, iterative pruning, and cluster summarization converge toward expert-aligned features. The gap between automated systems and expert knowledge is less about model capability than about how knowledge is elicited and refined. We release BrandGuide, the first dataset enabling systematic evaluation of expert alignment in feature engineering, and believe that grounding feature discovery in expert knowledge is a necessary step toward deploying ML in high stakes domains that demand human oversight. \looseness=-1

\textbf{Limitations.} FEST currently addresses binary classification; extension to multi-class settings requires architectural modifications. Feature quality is bounded by the underlying LLM's capabilities and biases. While substantially cheaper than Felix (86$\times$), FEST exceeds single-pass methods in cost. See Appendix~\ref{app:limitations} for detailed discussion.

\FloatBarrier
\bibliographystyle{plainnat}
\bibliography{paper}

\appendix

\section{Related Work}
\label{app:related_work}
\paragraph{1. Manual Feature Engineering.} Early AI and expert systems relied on manually crafted, human-interpretable features and rules. Classic examples include the MYCIN medical diagnosis system, which encoded clinical knowledge as explicit features and logical rules \citep{shortliffe2012computer}, and early spam filters that used hand-designed counts and lexical features to achieve practical performance \citep{sahami1998bayesian}. Manual feature design remains common in domains that demand transparency (e.g., medicine, law and regulated industries), but it is costly, slow, and limited by human cognitive biases and domain coverage \citep{kahneman2011thinking}. As data modalities and problem complexity increased, the scalability limits of purely manual engineering motivated a broad literature on automated and semi-automated feature construction.

\paragraph{2. Classical Automated Feature Engineering.} Classical automated feature engineering systems construct 
large candidate pools via predefined transformations and then select or prune useful features. Representative systems (e.g., 
AutoFeat and OpenFE) enumerate combinations, nonlinear transforms, or symbolic expressions and rely on statistical selection, 
regularization, or search procedures to find compact, predictive subsets \citep{horn2019autofeat, zhang2023openfe}. These 
approaches reduce human effort and can produce strong tabular baselines, but they generally assume an existing set of input 
columns (tabular format), depend on handcrafted transformation templates, and do not natively operate on raw unstructured inputs 
(e.g., images or free text). Consequently, classical pipelines transfer poorly to multimodal raw-data settings and offer limited 
semantic control over the meaning of constructed features.

\paragraph{3. LLM-Based Feature Engineering.} LLMs have recently been used as feature proposers in both structured and unstructured settings. 

\textit{Structured data.} LLMs have been applied to structured and tabular settings where 
feature transformations can be expressed programmatically. \citep{han2024large} propose FeatLLM, an in-context prompting approach 
that elicits interpretable rule-style features from an LLM for few-shot tabular tasks; generated candidates are evaluated with 
simple downstream learners and selected when useful.\citep{abhyankar2025llm} frame feature engineering as program search in 
LLM-FE, combining evolutionary optimization with LLM generation so that the model proposes programmatic transforms and a 
performance signal guides evolution of the feature population. \citet{batista2025embedding} study seeding classical pipelines with domain-aware LLM proposals to accelerate symbolic or evolutionary search, showing modest speedups and occasional accuracy gains from such knowledge injection. \citet{ko2025ferg} introduce FeRG-LLM, a fine-tuned LLM that uses iterative reasoning and local feedback to generate compact, locally interpretable features while remaining computationally efficient.

\textit{Unstructured data.} Recent work has explored using large language models as pattern miners and feature proposers for 
unstructured inputs. \citep{weiss2024exploration} documents how interactive prompting with ChatGPT can extract structured patterns 
and human-readable rules from heterogeneous data sources, illustrating a co-creative pattern-mining workflow that leverages LLM 
fluency for interpretability and hypothesis extraction. For text classification specifically, FELIX prompts LLMs to generate 
high-level textual descriptors or symbolic features from raw documents; these LLM-derived features feed lightweight classifiers 
and often outperform bag-of-words or raw embedding baselines while remaining human-interpretable \citep{malberg2024felix}. Such 
systems demonstrate that LLMs can surface semantically meaningful, deployable features from unstructured inputs, but many are 
implemented as one-shot or human-in-the-loop procedures rather than fully automated iterative pipelines.

These methods demonstrate that LLMs can propose semantically meaningful features, but most operate in a single generation pass, assume tabular inputs, or lack iterative refinement with downstream validation signals.

\paragraph{4. Concept Bottleneck Models.} Concept Bottleneck Models (CBMs) pursue a complementary goal: mapping raw inputs to human-interpretable intermediate concepts before prediction \citep{koh2020concept}. CBMs require predefined concept sets with labeled annotations, limiting scalability. Label-Free CBMs address this by using foundation models to automatically generate concept sets without labeled concept data, scaling to ImageNet \citep{oikarinen2023label}. TCAV provides a post-hoc alternative, using concept activation vectors to quantify concept importance in trained networks without modifying them \citep{kim2018interpretability}. FEST can be viewed as a label-free, iteratively refined concept bottleneck for unstructured data: it discovers the concept space from scratch via contrastive LLM prompting, deduplicates via semantic clustering, and validates via decision tree importance, without requiring predefined concept annotations or post-hoc probing.

\paragraph{5. Inherent Interpretability vs.\ Post-hoc Explanation.} \citet{rudin2019stop} argues that high-stakes domains should use inherently interpretable models rather than explaining black boxes post hoc, since post-hoc methods (LIME, SHAP, GradCAM) produce approximate attributions that may not faithfully reflect model reasoning \citep{ribeiro2016should, lundberg2017unified, selvaraju2017grad}. FEST aligns with this principle: its features are explicit, human-verifiable predicates (e.g., ``uses inclusive language,'' ``exclamation count'') that practitioners can audit before deployment, and its decision tree paths provide transparent classification logic. This contrasts with deep embedding approaches where interpretability requires auxiliary explanation tools.

\paragraph{6. Iterative Self-Refinement.} FEST's generate-evaluate-refine loop connects to the broader paradigm of iterative self-improvement in LLM systems. Self-Refine \citep{madaan2023self} demonstrated that LLMs can generate output, critique it, and iteratively improve, achieving 20\% average gains across diverse tasks. FEST applies an analogous loop to feature engineering: LLMs generate candidate features, decision trees evaluate their discriminative value, and weak features are pruned while new candidates are proposed in subsequent iterations. The key difference is that FEST's refinement signal comes from an external classifier (decision tree importance) rather than LLM self-critique, grounding the evolution in empirical task performance.

\paragraph{7. LLM-as-Judge Evaluation.} \citet{zheng2023judging} established that strong LLMs can approximate human judgment with $>$80\% agreement, enabling scalable evaluation of open-ended outputs. FEST adopts this paradigm for expert coverage evaluation: GPT-4o judges whether discovered features capture the same brand dimensions as expert guidelines, providing interpretable 0--10 scores that eliminate reliance on arbitrary embedding similarity thresholds.

\paragraph{8. Comparison with FEST:} While prior work demonstrates the promise of LLMs for feature discovery across both unstructured and structured settings, FEST departs from these approaches along several dimensions. First, FEST targets raw multimodal observational data (text, images, and tabular inputs) by prompting LLMs to generate \emph{interpretable} feature candidates directly from example pairs rather than requiring an initial tabular feature bank. Second, rather than performing a single stage of proposal and selection, FEST implements a \emph{self-evolving} generate, deduplicate, validate loop: (i) LLMs propose diverse semantic and deterministic features from pairwise comparisons, (ii) semantic embeddings and clustering compress paraphrastic or duplicate proposals, and (iii) decision trees evaluate, provide feature-importance signals, and guide iterative pruning and feature-bank evolution. This closed-loop design (including the use of probabilistic LLM feature inference scores and a feature importance history for robust pruning) enables continuous refinement and avoids the redundancy and spuriousness typical of one-shot generators. Third, FEST explicitly prioritizes semantic control and human interpretability: deduplicated feature summaries and tree decision paths provide transparent, human-readable logic that practitioners can inspect and adjust, contrasting with approaches that rely primarily on opaque performance signals or implicit model internals \citep{han2024large,abhyankar2025llm,malberg2024felix}. Finally, FEST’s pairwise comparison strategy and tree-based validation make it robust to common-attribute noise and permit recovery of compositional logical relationships (as validated on controlled benchmarks), positioning FEST as a more general, controllable, and interpretable automated feature engineering methodology for both raw and structured data.

\paragraph{9. Connection to Robust Optimization.} FEST shares a conceptual affinity with Invariant Risk Minimization (IRM) \citep{arjovsky2019invariant} and Group DRO \citep{sagawa2019distributionally}: both seek features stable across environments rather than spuriously correlated with labels. FEST's iterative pruning of low-importance features echoes invariance-seeking, as features that do not generalize across batches are progressively eliminated. The key distinction is that IRM and Group DRO operate on a \textit{fixed} feature space with explicit environment labels, whereas FEST \textit{constructs} the feature space from raw unstructured data. FEST thus addresses a problem upstream of robust optimization: discovering interpretable features in the first place. These are complementary; robust optimization could be applied atop FEST-discovered features for further out-of-distribution gains.

\section{Algorithm Pseudocode}
Algorithm~\ref{algo:algo} presents the complete FEST pseudocode.

\begin{algorithm}
    \caption{FEST: Feature Engineering with Self-evolving Trees}
    \label{algo:algo}
    \begin{algorithmic}[1]
    \small
    \STATE \textbf{Input:} Dataset $D$ with binary labels, hyperparameters $\tau_{accuracy}$, $\tau_{importance}$, $K$ (batch size)
    \STATE \textbf{Output:} Feature bank $F$ and trained decision tree $\mathcal{T}$
    \STATE \textbf{Initialize:}
    \STATE $F \gets \emptyset$ \COMMENT{Global feature bank (or initialize with expert features if available)}
    \STATE $\mathcal{P} \gets$ ConstructPairs($D$) \COMMENT{Create comparison pairs $(x^+, x^-)$}
    \STATE $\mathcal{P}_{train}, \mathcal{P}_{val}, \mathcal{P}_{test} \gets$ Split($\mathcal{P}$) \COMMENT{Train/validation/test split}
    \STATE $H_{importance} \gets \emptyset$ \COMMENT{Feature importance history}
    \STATE $t \gets 0$ \COMMENT{Iteration counter}
    \STATE
    \WHILE{validation accuracy $< \tau_{accuracy}$ and pairs remain}
        \STATE $t \gets t + 1$
        \STATE $B \gets$ NextBatch($\mathcal{P}_{train}$, $K$) \COMMENT{Get batch of $|B|$ pairs}
        \STATE
        \STATE \textbf{// Stage 1: Dual-Stream Feature Generation}
        \STATE \textbf{/* Semantic Feature Generation */}
        \STATE $F_{SE}^{(t)} \gets \emptyset$
        \FOR{each pair $(x_i^+, x_i^-) \in B$}
            \FOR{each prompt template $p \in \{1, \ldots, M\}$}
                \STATE $f \gets$ LLM.GenerateSEFeature($x_i^+, x_i^-$, template $p$) \COMMENT{Pairwise comparison}
                \STATE $F_{SE}^{(t)} \gets F_{SE}^{(t)} \cup \{f\}$
            \ENDFOR
        \ENDFOR
        \STATE
        \STATE \textbf{/* Deterministic Feature Generation */}
        \STATE $B_{sample} \gets$ Sample($B$, size=5) \COMMENT{Sample few pairs for task-level generation}
        \STATE $F_{DE}^{(t)} \gets$ LLM.GenerateDEFeatures($B_{sample}$, task\_desc, $F$) \COMMENT{Task-level, returns Python functions}
        \STATE $F_{DE}^{(t)} \gets$ ValidateAndRefine($F_{DE}^{(t)}$) \COMMENT{Execute in sandbox, fix errors}
        \STATE
        \STATE \textbf{// Stage 2: Semantic Deduplication and Bank Merging}
        \STATE $E_{SE} \gets$ ConditionalEmbedding($F_{SE}^{(t)}$, task\_desc) \COMMENT{Domain-aware embeddings}
        \STATE $C_{clusters} \gets$ KMeansClustering($E_{SE}$) \COMMENT{Cluster semantically similar SE features}
        \STATE $\bar{F}_{SE}^{(t)} \gets$ LLM.SummarizeClusters($C_{clusters}$) \COMMENT{One representative per cluster}
        \STATE $\bar{F}_{SE}^{(t)} \gets$ SemanticMerge($\bar{F}_{SE}^{(t)}$, $F$) \COMMENT{Remove duplicates with existing bank}
        \STATE $F^{(t)} \gets F \cup \bar{F}_{SE}^{(t)} \cup F_{DE}^{(t)}$ \COMMENT{Current iteration's complete feature set}
        \STATE
        \STATE \textbf{// Stage 3: Feature Inference and Encoding}
        \STATE Split pairs in $B$ into individual samples: $S \gets \{x_1^+, x_1^-, \ldots, x_{|B|}^+, x_{|B|}^-\}$ with $|S| = 2|B|$
        \STATE
        \FOR{each sample $x_i \in S$ and SE feature $f_j \in F_{SE}$ (all semantic features in $F^{(t)}$)}
            \STATE $p_1, p_0 \gets$ LLM.GetProbs($f_j$, $x_i$) \COMMENT{Probabilities for tokens "1" and "0"}
            \STATE $X_{SE,ij} \gets g_{SE}(f_j, x_i) = \frac{p_1}{p_1 + p_0}$ \COMMENT{Normalized confidence}
        \ENDFOR
        \STATE
        \FOR{each sample $x_i \in S$ and DE feature $f_j \in F_{DE}$ (all deterministic features in $F^{(t)}$)}
            \STATE $X_{DE,ij} \gets g_{DE}(f_j, x_i) = \texttt{function}_j(x_i)$ \COMMENT{Execute Python function}
        \ENDFOR
        \STATE
        \STATE $\mathbf{X} \gets [\mathbf{X}_{SE} \mid \mathbf{X}_{DE}]$ \COMMENT{Concatenate: $\mathbf{X} \in \mathbb{R}^{2|B| \times (|F_{SE}| + |F_{DE}|)}$}
        \STATE $\mathbf{y} \gets [1, 0, \ldots, 1, 0]^\top$ \COMMENT{Labels: 1 for $x^+$, 0 for $x^-$, $\mathbf{y} \in \{0,1\}^{2|B|}$}
        \STATE
        \STATE \textbf{// Stage 4: Tree Training and Evolution}
        \STATE $\mathcal{T} \gets$ DecisionTree.Train($\mathbf{X}, \mathbf{y}$) \COMMENT{Train decision tree}
        \STATE $acc_{val} \gets$ Evaluate($\mathcal{T}$, $\mathcal{P}_{val}$, $F^{(t)}$) \COMMENT{Validation accuracy}
        \IF{$acc_{val} \geq \tau_{accuracy}$}
            \STATE \textbf{break} \COMMENT{Convergence achieved}
        \ENDIF
        \STATE
        \STATE $\mathbf{I}^{(t)} \gets$ $\mathcal{T}$.GetFeatureImportance() \COMMENT{Extract importance scores}
        \STATE $H_{importance} \gets$ UpdateHistory($H_{importance}$, $\mathbf{I}^{(t)}$, $F^{(t)}$)
        \STATE $F^{(t)'} \gets$ Prune($F^{(t)}$, $H_{importance}$, $\tau_{importance}$) \COMMENT{Remove low-importance features}
        \STATE $F \gets F^{(t)'}$ \COMMENT{Update global bank for next iteration}
    \ENDWHILE
    \STATE
    \STATE \textbf{return} $F$, $\mathcal{T}$
    \end{algorithmic}
\end{algorithm}

\section{Feature Encoding Details}
\label{app:encoding_math}

This section provides the full mathematical formulation of feature encoding summarized in \S\ref{sec:feature_encoding}.

\textbf{Semantic Features Encoding.} For each individual sample $x$ and each semantic feature $f_k \in F_{SE}$, we prompt the LLM to evaluate whether $f_k$ is present or absent. Rather than using binary responses, we extract richer signal from LLM uncertainty: we obtain output probabilities $p_\theta(y = 1 \mid f_k, x)$ and $p_\theta(y = 0 \mid f_k, x)$ for tokens ``1'' (present) and ``0'' (absent), then compute normalized confidence:

\begin{equation}
g_{SE}(f_k, x)
\;=\;
\frac{
p_\theta(y = 1 \mid f_k, x)
}{
p_\theta(y = 1 \mid f_k, x)
+
p_\theta(y = 0 \mid f_k, x)
}
\end{equation}

where $p_\theta$ denotes the LLM parameterized by $\theta$, and $y \in \{0,1\}$ represents the prediction. This confidence score in $[0, 1]$ captures LLM certainty about feature presence, enabling more nuanced encoding than binary labels. We construct the semantic feature matrix $\mathbf{X}_{SE} \in \mathbb{R}^{2|B| \times |F_{SE}|}$ where each element $X_{SE,ij} = g_{SE}(f_j, x_i)$.

\textbf{Deterministic Features Encoding.} For each sample $x$ and each deterministic feature $f_k \in F_{DE}$, we execute the associated Python function: $g_{DE}(f_k, x) = \texttt{function}_k(x)$, returning numeric values. We construct the deterministic feature matrix $\mathbf{X}_{DE} \in \mathbb{R}^{2|B| \times |F_{DE}|}$ where each element $X_{DE,ij} = g_{DE}(f_j, x_i)$.

\textbf{Combined Feature Matrix.} We concatenate the semantic and deterministic feature matrices to obtain the final feature representation: $\mathbf{X} = [\mathbf{X}_{SE} \mid \mathbf{X}_{DE}] \in \mathbb{R}^{2|B| \times (|F_{SE}| + |F_{DE}|)}$. The corresponding label vector is $\mathbf{y} \in \{0,1\}^{2|B|}$ where $y_i = 1$ for positive class samples and $y_i = 0$ for negative class samples.

\section{Pairwise Comparison Motivation}
\label{app:pairwise_motivation}

Absolute assessment of individual samples suffers from two fundamental limitations: (1) inability to distinguish universally present attributes from discriminative features, and (2) dependence on subjective absolute thresholds. For example, in news headline analysis, absolute assessment might identify ``contains numbers'' as relevant without recognizing that numbers appear equally in both successful and unsuccessful headlines. Pairwise comparison identifies features that actually differentiate: successful headlines use specific question formats while unsuccessful ones use generic statements. This design is supported by recent work \citep{geng2025delta} showing that the quality delta between samples in a pair provides a richer learning signal than individual samples. FEST's convergence and pruning thresholds ($\tau_{accuracy}$, $\tau_{importance}$) are algorithmic flow-control parameters, distinct from the semantic feature-definition thresholds criticized in absolute assessment approaches.

\section{Additional Results}
\label{appendix:additional_results}

\begin{itemize}[leftmargin=8pt]
    \item Brand classification (text) results in Table~\ref{tab:brandwise_classification_results_text}
    \item Brand classification (image) results in Table~\ref{tab:brandwise_classification_results_image}
\end{itemize}

\begin{table}[!tb]
\centering
\caption{\small{Brand-wise classification accuracy (\%) for text-based brand validation across 5 brands (Emirates, Adobe, Porsche, Louis Vuitton, Pizza Hut) and 3 classifiers (DT, LR, RF). Main text Table~\ref{tab:classification_results} reports averages across brands. Bold indicates best per brand-classifier combination.}}
\label{tab:brandwise_classification_results_text}
\resizebox{0.8\linewidth}{!}{%
\begin{tabular}{ccccccc}
\toprule
\multirow{2}{*}{\textbf{Classifier}} &
\multirow{2}{*}{\textbf{Feature Generator}} &
\multicolumn{5}{c}{\textbf{Brand}} \\ \cline{3-7}
 & & Emirates & Adobe & Porsche & Louis Vuitton & Pizza Hut\\ \toprule

\multirow{5}{*}{DT}
 & Zero-Shot LLM     & 69.60 & 72.80 & 60.40 & 85.60 & 78.05\\ %
 & Few-Shot LLM     & 69.20 & 81.73 & 68.93 & \best{86.53} & 75.20 \\
 & Felix & 67.60 & 69.60 & 63.20 & 80.00 & \best{81.09} \\
 & \fc FEST (Ours) & \fc\textbf{73.60} & \fc\textbf{88.40} & \fc\textbf{77.20} & \fc 83.60 & \fc 78.05 \\ \midrule

\multirow{5}{*}{LR}
 & Zero-Shot LLM     &  64.40 & 66.40 & 68.00 & 78.40 & 70.00 \\
 & Few-Shot LLM     &   61.06 & 82.80 & 64.00 & 87.06 & 69.91\\
 & Felix & 72.00 & 77.20 & 66.80 & 80.40 & 82.92 \\
 & \fc FEST (Ours) & \fc\textbf{73.60} & \fc\textbf{86.00} & \fc\textbf{76.80} & \fc\textbf{88.00} & \fc\textbf{84.14} \\ \midrule

\multirow{5}{*}{RF}
 & Zero-Shot LLM     &  80.40 & 76.40 & 71.20 & 88.00 & 85.37 \\
 & Few-Shot LLM     & 77.99 & \best{88.80} & 75.73 & 89.60 & 83.73\\
 & Felix & 76.80 & 81.60 & 71.60 & 84.80 & 85.36 \\
 & \fc FEST (Ours) & \fc\textbf{82.80} & \fc 88.00 & \fc\textbf{78.80} & \fc\textbf{90.00} & \fc\textbf{85.97} \\ \midrule

 \multirow{5}{*}{MLP}
 & Zero-Shot LLM     &  69.20 & 76.80 & 69.60 & 86.80 &  77.43 \\
 & Few-Shot LLM     &   72.00 & 82.53 & 67.33 & 87.73 & 74.18\\
 & Felix & 72.40 & 77.20 & \best{74.00} & 84.80 & \best{82.92} \\
 & \fc FEST (Ours) & \fc\textbf{78.80} & \fc\textbf{85.20} & \fc 72.00 & \fc\textbf{90.80} & \fc 81.09 \\ \midrule

 \multirow{5}{*}{XGB}
 & Zero-Shot LLM     &  78.40 & 75.20 & 70.80 & 87.60 & 83.53 \\
 & Few-Shot LLM     &   77.86 & 87.46 & 73.30 & 89.40 & \best{87.19}\\
 & Felix & 78.00 & 80.80 & 74.00 & 84.00 & 86.58 \\
 & \fc FEST (Ours) & \fc\textbf{83.60} & \fc\textbf{88.00} & \fc\textbf{78.40} & \fc\textbf{90.00} & \fc 82.92 \\ \midrule
\end{tabular}}
\end{table}

\begin{table}[!tb]
\centering
\caption{\small{Brand-wise classification accuracy (\%) for image-based brand validation across 5 brands and 3 classifiers (DT, LR, RF). Main text Table~\ref{tab:classification_results} reports averages across brands. Bold indicates best per brand-classifier combination.}}
\label{tab:brandwise_classification_results_image}
\resizebox{0.8\linewidth}{!}{%
\begin{tabular}{ccccccc}
\toprule
\multirow{2}{*}{\textbf{Classifier}} &
\multirow{2}{*}{\textbf{Feature Generator}} &
\multicolumn{5}{c}{\textbf{Brand}} \\ \cline{3-7}
 & & Emirates & Adobe & Porsche & Louis Vuitton & Pizza Hut\\ \toprule

\multirow{4}{*}{DT}
 & Zero-Shot LLM     & 60.49 & 67.33 & 58.40 & 62.70 & 77.16\\
 & Few-Shot LLM     & 66.27 & 75.50 & 64.80 & 66.40 & 83.53 \\
 & Felix & 60.40 & 77.20 & 58.00 & 58.80 & 84.70 \\
 & \fc FEST (Ours) & \fc\best{76.80} & \fc\best{82.40} & \fc\best{71.20} & \fc\best{72.80} & \fc\best{88.41} \\ \midrule

\multirow{4}{*}{LR}
 & Zero-Shot LLM     &  62.96 & 74.59 & 67.20 & 70.08 & 77.16 \\
 & Few-Shot LLM     &   62.65 & 78.31 & 68.40 & 72.40 & 83.53\\
 & Felix & 62.40 & \best{82.40} & 60.80 & 66.80 & 79.26 \\
 & \fc FEST (Ours) & \fc\best{75.60} & \fc 81.20 & \fc\best{68.80} & \fc\best{72.80} & \fc\best{89.02} \\ \midrule

\multirow{4}{*}{RF}
 & Zero-Shot LLM     &  64.60 & 79.43 & 70.80 & 74.18 & 83.33 \\
 & Few-Shot LLM     & 75.10 & 83.94 & 74.80 & 75.60 & 84.14\\
 & Felix & 59.60 & 83.60 & 59.60 & 64.80 & 85.30 \\
 & \fc FEST (Ours) & \fc\best{81.60} & \fc\best{85.60} & \fc\best{70.40} & \fc\best{78.70} & \fc\best{91.46} \\ \midrule

\multirow{4}{*}{MLP}
 & Zero-Shot LLM     &  67.90 & 74.19 & 66.40 & 68.85 & 74.07 \\
 & Few-Shot LLM     &   70.28 & 73.09 & \best{70.40} & 70.80 & 82.90\\
 & Felix & 63.20 & \best{80.80} & 57.60 & 58.40 & 84.14 \\
 & \fc FEST (Ours) & \fc\best{78.00} & \fc 79.20 & \fc 64.80 & \fc\best{71.20} & \fc\best{89.63} \\ \midrule

\multirow{4}{*}{XGB}
 & Zero-Shot LLM     &  65.02 & 77.01 & 68.40 & 72.54 & 82.17 \\
 & Few-Shot LLM     &   70.28 & 80.32 & \best{74.80} & 74.80 & 84.14\\
 & Felix & 59.20 & 84.00 & 60.80 & 67.20 & 87.19 \\
 & \fc FEST (Ours) & \fc\best{82.00} & \fc\best{84.40} & \fc 71.20 & \fc\best{78.00} & \fc\best{90.85} \\ \bottomrule
\end{tabular}}
\end{table}

\section{Limitations}
\label{app:limitations}

While FEST demonstrates effective feature discovery and refinement capabilities, several limitations warrant discussion:

\textbf{Binary classification scope}: FEST currently addresses binary classification tasks. Extension to multi-class classification and regression requires architectural modifications, particularly in pairwise comparison formulation and decision tree feedback mechanisms.

\textbf{Incomplete expert feature coverage}: FEST achieves 60--80\% coverage of expert-designed features across brand classification tasks (measured via LLM-as-judge evaluation). An audit of uncovered guidelines reveals they are structurally unobservable from post text (organizational policies, abstract brand philosophy, platform meta-guidelines). Achieving higher coverage may require multi-modal inputs or interactive expert feedback during feature generation.

\textbf{LLM-dependent feature quality}: Feature quality is bounded by the underlying LLM's capabilities and biases. While semantic deduplication and tree-guided pruning mitigate spurious features, the framework inherits limitations of the base LLM. On tasks where semantic features suffice (e.g., stress detection), simpler approaches may achieve comparable performance.

\textbf{Computational overhead}: Iterative feature generation, semantic clustering, and model retraining impose computational costs exceeding single-pass methods. For brand classification, FEST requires approximately 16 minutes per task on average (\$0.10 per run), compared to Felix's 69 minutes (\$8.62 per run). While 86$\times$ cheaper than the nearest baseline, these costs are amortized over deployment and may limit real-time applications. Inference requires only a single LLM pass against 10--15 retained features.

\textbf{Correlation versus causation}: Discovered features represent predictive patterns, not causal relationships. While decision tree paths provide interpretable rules, practitioners must validate features against domain knowledge before deployment in high-stakes contexts. Integration with causal discovery methods remains future work.

\textbf{Limited domain validation}: Evaluation focuses on marketing, content moderation, and psychological assessment tasks. Validation in critical domains (healthcare diagnosis, financial fraud detection, legal decision-making) requiring stricter safety and regulatory compliance remains necessary before broader adoption.

Overall, these limitations point to directions for future work, including bias mitigation in LLM feature generation, integration with causal discovery methods and optimization for large-scale deployment.

\section{Impact Statement}
\label{sec:impact}
FEST's automation of feature engineering for high-stakes domains carries both benefits and mitigated risks. By democratizing access to expert-validated feature discovery, FEST can accelerate ML deployment in domains like marketing, healthcare, and legal decision-making where manual feature engineering currently limits scalability. The framework's interpretability through practitioner-inspectable features and transparent decision tree paths supports accountability and enables domain experts to verify automated discoveries before deployment.

Importantly, FEST's demonstrated ability to discover features relevant to domain experts (as validated by 60-80\% coverage of brand voice characteristics) provides empirical validation that mitigates risks typically associated with fully automated methods. The expert refinement capability further aligns discovered features with domain knowledge, reducing concerns about arbitrary or opaque feature generation. However, practitioners should maintain human oversight to prevent perpetuation of biases present in training data or expert guidelines. The framework should augment rather than replace domain expertise, particularly in contexts where erroneous predictions carry significant consequences. We recommend validation of FEST-discovered features by domain experts before production deployment in sensitive domains.

\section{Future Work}

Beyond improving the scalability and causal grounding of FEST, we envision several exciting directions where the framework could extend its impact.

\textbf{Content optimization through actionable feedback}: Beyond classification, FEST generates explicit, interpretable features that form the decision path for each prediction. This opens the door to applications such as optimizing headlines, tweets, or advertisements. For example, when distinguishing between engaging and non-engaging content, FEST can surface the exact linguistic or structural attributes that influence predicted engagement. Practitioners can then receive concrete feedback such as ``headline length too short" or ``absence of emotional keywords", allowing them to modify content in ways directly aligned with model logic. This shifts predictive modeling from passive forecasting toward active guidance.

\textbf{Post-hoc explainability of black-box models}: Another intriguing direction is to repurpose FEST as an explanation layer for opaque models. Suppose a neural network achieves state-of-the-art accuracy on a classification task. By labeling data with the network’s predictions and then running FEST on top, one can extract interpretable features and decision rules that approximate the network’s learned representations. This would combine the high performance of black-box models with FEST’s ability to articulate insights in natural language, offering practitioners a window into otherwise inscrutable models.

\textbf{Scalability and causal discovery}: On the methodological side, future work should push FEST toward more efficient large-scale deployment and explore causal discovery. Enhancing the efficiency of the generate, deduplicate, validate loop will make FEST practical for massive datasets and near real-time applications. Integrating causal reasoning into the feature refinement process could help distinguish predictive correlations from genuine drivers of outcomes, a particularly critical need in scientific and policy domains.

Taken together, these directions suggest that FEST is not only a framework for automating feature engineering but also a step toward rethinking the role of models in human decision-making: from opaque predictors to transparent copilots that explain, advise and guide.

\section{Ablation: Semantic-only / Deterministic-only / SE+DE}
\label{app:ablation_sede}

We run FEST in three configurations across 7 tasks (5 brand text classification tasks + content authenticity detection + stress detection) to isolate the contribution of each feature stream.

\begin{table}[H]
\centering
\small
\caption{\small{Dual-stream ablation (accuracy \%) with Decision Tree and Random Forest classifiers across 7 tasks (5 brand text classification + content authenticity + stress detection). SE+DE (full FEST) outperforms both single-stream variants in 11 of 14 task-classifier combinations. The three exceptions (Louis Vuitton DT, Adobe RF, Porsche RF) show SE-only marginally outperforming by 0.4--3.2pp, indicating DE features add value in most settings and do not hurt in the rest. DE-only is consistently the weakest, confirming semantic features form the core signal while deterministic features add measurable precision. Bold indicates best per task-classifier.}}
\begin{tabular}{lcccccc}
\toprule
& \multicolumn{3}{c}{\textbf{Decision Tree}} & \multicolumn{3}{c}{\textbf{Random Forest}} \\
\textbf{Task} & \textbf{DE} & \textbf{SE} & \textbf{SE+DE} & \textbf{DE} & \textbf{SE} & \textbf{SE+DE} \\ \toprule
Emirates & 67.2 & 72.8 & \textbf{73.6} & 70.8 & 76.4 & \textbf{82.8} \\
Adobe & 68.0 & 83.2 & \textbf{88.4} & 74.8 & \textbf{90.4} & 88.0 \\
Porsche & 72.8 & 71.2 & \textbf{77.2} & 74.0 & \textbf{79.2} & 78.8 \\
Louis Vuitton & 79.2 & \textbf{86.8} & 83.6 & 84.8 & 89.8 & \textbf{90.0} \\
Pizza Hut & 74.8 & 72.0 & \textbf{78.1} & 79.2 & 78.0 & \textbf{86.0} \\ \hline
Content Authenticity & 63.2 & 87.6 & \textbf{91.2} & 91.2 & 94.4 & \textbf{97.2} \\ \hline
Stress Detection & 58.4 & 72.0 & \textbf{78.0} & 63.2 & 78.8 & \textbf{83.6} \\ \bottomrule
\end{tabular}
\end{table}

SE+DE wins in 11 of 14 DT+RF combinations. The three exceptions (Louis Vuitton DT, Adobe RF, Porsche RF) show SE-only marginally outperforming by 0.4--3.2pp, indicating that DE features add value in most settings and do not hurt in the rest. DE-only is always the weakest, confirming that semantic features form the core signal while deterministic features contribute complementary measurable discriminative power.

\section{Ablation: Expert Refinement Disentanglement}
\label{app:abc_ablation}

To disentangle the contributions of refinement and augmentation in expert feature refinement, we evaluate three conditions for brand text classification:
\begin{itemize}[leftmargin=8pt]
    \item \textbf{A (Expert alone)}: Expert-designed features used as-is.
    \item \textbf{B (Expert Refined)}: FEST features filtered to those semantically similar to expert features (cosine $\geq 0.7$), capturing refinement without augmentation.
    \item \textbf{C (Expert+FEST)}: Full FEST output including augmented features.
\end{itemize}

Table~\ref{tab:expert_refinement} presents the full per-classifier breakdown. B$>$A in most cases demonstrates that refinement alone improves over static expert features. C$>$B in all cases demonstrates that augmentation provides further independent gains. The one exception (LR Adobe: B=78.4 $<$ A=80.4) may reflect LR sensitivity to feature set size changes, but C still wins by +8.0pp.

\begin{table}[H]
\centering
\caption{\small{Expert feature refinement accuracy (\%) for text brand classification across three brands (Adobe, LG, Porsche) and four classifiers. Column shading encodes the performance tier: \cbronze~\textbf{A (bronze)} uses expert-authored features as-is; \csilver~\textbf{B (silver)} replaces them with FEST-discovered features semantically aligned to expert guidelines (LLM-judge score $\geq 7$), isolating the effect of refinement without augmentation; \cgold~\textbf{C (gold)} adds all remaining FEST features (Expert+FEST), measuring the further benefit of augmentation. B$>$A in most cases confirms that FEST operationalizes ambiguous guidelines into more discriminative definitions; C$>$B in all cases confirms that augmented features provide complementary gains. Gain (C$-$A) is reported in percentage points. Avg rows (bold) average over DT, LR, RF, and LLM classifiers.}}
\label{tab:expert_refinement}
\setlength{\tabcolsep}{6pt}
\resizebox{0.8\linewidth}{!}{%
\begin{tabular}{clcccc}
\toprule
\textbf{Clf.} & \textbf{Brand} & \textbf{\cbronze A (Expert)} & \textbf{\csilver B (Refined)} & \textbf{\cgold C (Exp+FEST)} & \textbf{Gain (C$-$A)} \\ \midrule
\multirow{3}{*}{DT} & Adobe & \cbronze 78.40 & \csilver 87.20 & \cgold 88.40 & \textcolor{green!50!black}{+10.00} \\
& LG & \cbronze 69.60 & \csilver 74.15 & \cgold 75.42 & \textcolor{green!50!black}{+5.82} \\
& Porsche & \cbronze 82.00 & \csilver 84.40 & \cgold 88.80 & \textcolor{green!50!black}{+6.80} \\ \midrule
\multirow{3}{*}{LR} & Adobe & \cbronze 80.40 & \csilver 78.80 & \cgold 90.40 & \textcolor{green!50!black}{+10.00} \\
& LG & \cbronze 69.20 & \csilver 75.42 & \cgold 83.47 & \textcolor{green!50!black}{+14.27} \\
& Porsche & \cbronze 77.60 & \csilver 86.40 & \cgold 90.00 & \textcolor{green!50!black}{+12.40} \\ \midrule
\multirow{3}{*}{RF} & Adobe & \cbronze 84.40 & \csilver 91.20 & \cgold 92.00 & \textcolor{green!50!black}{+7.60} \\
& LG & \cbronze 79.60 & \csilver 82.62 & \cgold 86.44 & \textcolor{green!50!black}{+6.84} \\
& Porsche & \cbronze 85.20 & \csilver 86.40 & \cgold 88.40 & \textcolor{green!50!black}{+3.20} \\ \midrule
\multirow{3}{*}{LLM} & Adobe & \cbronze 54.40 & \csilver 56.00 & \cgold 73.60 & \textcolor{green!50!black}{+19.20} \\
& LG & \cbronze 53.38 & \csilver 54.66 & \cgold 58.05 & \textcolor{green!50!black}{+4.67} \\
& Porsche & \cbronze 71.20 & \csilver 73.20 & \cgold 73.60 & \textcolor{green!50!black}{+2.40} \\ \midrule
\multirow{3}{*}{\textbf{Avg}} & Adobe & \cbronze\textbf{74.40} & \csilver\textbf{78.30} & \cgold\textbf{86.10} & \textcolor{green!50!black}{\textbf{+11.70}} \\
& LG & \cbronze\textbf{67.94} & \csilver\textbf{71.71} & \cgold\textbf{75.84} & \textcolor{green!50!black}{\textbf{+7.90}} \\
& Porsche & \cbronze\textbf{79.00} & \csilver\textbf{82.60} & \cgold\textbf{85.20} & \textcolor{green!50!black}{\textbf{+6.20}} \\ \bottomrule
\end{tabular}}
\end{table}

\section{Feature Evolution Analysis}
\label{app:feature_evolution}

\begin{figure}[H]
\centering
\includegraphics[width=\textwidth]{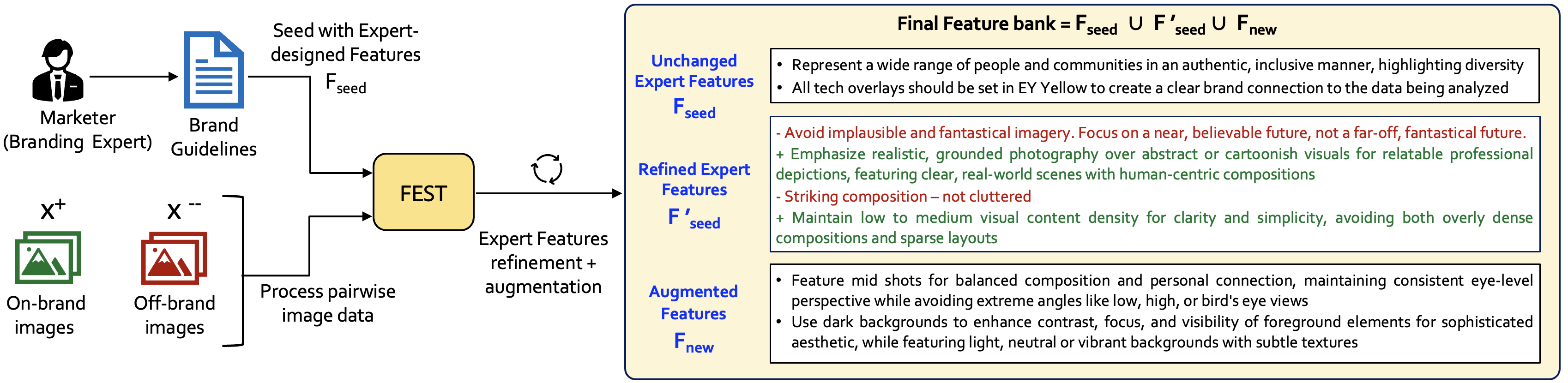}
\caption{\small{FEST refines and augments expert features (EY imagery). From seeds $F_{seed}$, FEST produces unchanged, refined ($F'_{seed}$), and augmented ($F_{new}$) features.}}
\label{fig:expert_refinement}
\end{figure}

Figure~\ref{fig:expert_refinement} visualizes the feature bank evolution for EY brand image classification. The final features fall into three groups:

\begin{enumerate}[leftmargin=16pt]
\item \textit{Unchanged Expert Features} ($F_{seed}$): features retained verbatim because they are already precise and actionable (e.g., ``All tech overlays should be set in EY Yellow to create a clear brand connection'').

\item \textit{Refined Expert Features} ($F'_{seed}$): ambiguous guidelines transformed into precise operational definitions. For example, ``Avoid implausible and fantastical imagery. Avoid unoriginal or unsophisticated imagery\ldots'' was refined to ``Emphasize realistic, grounded photography over abstract or cartoonish visuals for relatable professional depictions, featuring clear, real-world scenes with human-centric compositions.'' Similarly, Porsche's text guideline ``Narrative told through the voice of people and their personal stories'' was operationalized as ``Emphasize emotional storytelling and personal connections over technical details and promotional language.'' These refinements convert implicit expert knowledge into explicit, measurable criteria that LLMs can reliably apply.

\item \textit{Augmented Features} ($F_{new}$): novel features discovered by FEST that extend beyond documented guidelines, surfacing implicit domain knowledge (e.g., ``Feature mid shots for balanced composition and personal connection, maintaining consistent eye-level perspective while avoiding extreme angles'' for EY images).
\end{enumerate}

\section{Variance Analysis}
\label{app:variance}

We run 3 independent seeds on brand text classification (5 brands $\times$ 5 classifiers = 25 combinations).

\begin{table}[H]
\centering
\small
\caption{\small{FEST stability: mean $\pm$ std accuracy (\%) over 3 independent seeds across all tasks. For brand text classification, the maximum std is 3.03pp (MLP/Emirates, XGB/Porsche), with the majority below 2pp. Brand image classification shows higher variance (max 6.2pp, XGB/Porsche), reflecting greater stochasticity in visual feature generation. Content authenticity and stress detection also exhibit larger variance for LR and MLP classifiers (up to 6.9pp), likely due to smaller dataset sizes amplifying seed sensitivity. K-means deduplication and tree-based pruning stabilize stochastic LLM generation into consistent feature banks; residual variance stems from classifier sensitivity rather than feature bank instability.}}
\begin{tabular}{llccccc}
\toprule
\textbf{Task} & \textbf{Dataset} & \textbf{DT} & \textbf{LR} & \textbf{RF} & \textbf{MLP} & \textbf{XGB} \\ \toprule
\multirow{6}{*}{\makecell[l]{Brand\\Classification\\(Text)}}
& \textit{Average} & 79.9 & 81.6 & 84.6 & 83.5 & 84.9 \\ \cline{2-7}
& Adobe & 87.3{\tiny$\pm$1.1} & 86.0{\tiny$\pm$0.9} & 88.7{\tiny$\pm$0.9} & 86.7{\tiny$\pm$0.8} & 89.1{\tiny$\pm$0.4} \\
& Emirates & 73.6{\tiny$\pm$2.1} & 73.7{\tiny$\pm$2.0} & 80.7{\tiny$\pm$3.0} & 78.5{\tiny$\pm$3.0} & 81.1{\tiny$\pm$1.9} \\
& Porsche & 76.0{\tiny$\pm$1.4} & 76.3{\tiny$\pm$1.2} & 78.4{\tiny$\pm$0.9} & 76.4{\tiny$\pm$1.4} & 79.5{\tiny$\pm$3.0} \\
& Louis Vuitton & 84.7{\tiny$\pm$2.5} & 87.6{\tiny$\pm$0.8} & 89.9{\tiny$\pm$0.2} & 89.9{\tiny$\pm$0.8} & 90.5{\tiny$\pm$0.5} \\
& Pizza Hut & 78.1{\tiny$\pm$1.1} & 84.5{\tiny$\pm$0.9} & 85.4{\tiny$\pm$0.9} & 86.2{\tiny$\pm$1.9} & 84.4{\tiny$\pm$1.3} \\ \midrule
\multirow{6}{*}{\makecell[l]{Brand\\Classification\\(Images)}}
& \textit{Average} & 78.9 & 79.1 & 84.6 & 78.6 & 84.6 \\ \cline{2-7}
& Adobe & 81.9{\tiny$\pm$0.4} & 83.7{\tiny$\pm$1.9} & 87.6{\tiny$\pm$0.5} & 81.1{\tiny$\pm$1.9} & 86.5{\tiny$\pm$1.5} \\
& Emirates & 77.9{\tiny$\pm$1.2} & 77.6{\tiny$\pm$1.4} & 83.7{\tiny$\pm$1.5} & 77.6{\tiny$\pm$0.3} & 83.5{\tiny$\pm$1.1} \\
& Porsche & 72.7{\tiny$\pm$2.2} & 73.0{\tiny$\pm$3.0} & 77.0{\tiny$\pm$4.8} & 70.8{\tiny$\pm$4.4} & 79.3{\tiny$\pm$6.2} \\
& Louis Vuitton & 73.0{\tiny$\pm$2.4} & 73.3{\tiny$\pm$1.0} & 81.6{\tiny$\pm$2.5} & 74.5{\tiny$\pm$2.4} & 81.6{\tiny$\pm$3.3} \\
& Pizza Hut & 88.8{\tiny$\pm$3.5} & 88.0{\tiny$\pm$4.8} & 92.9{\tiny$\pm$2.5} & 89.0{\tiny$\pm$1.3} & 91.9{\tiny$\pm$3.3} \\ \midrule
\makecell[l]{Content Authenticity} & & 90.13{\tiny$\pm$0.7} & 79.47{\tiny$\pm$6.4} & 95.2{\tiny$\pm$1.8} & 80.1{\tiny$\pm$6.9} & 94.67{\tiny$\pm$1.0} \\ \midrule
\makecell[l]{Stress Detection} & & 77.6{\tiny$\pm$0.3} & 74.1{\tiny$\pm$4.6} & 82.1{\tiny$\pm$1.8} & 73.6{\tiny$\pm$5.6} & 80.5{\tiny$\pm$0.7} \\ \bottomrule
\end{tabular}
\end{table}

\section{Runtime and Cost Analysis}
\label{app:runtime}

\begin{table}[H]
\centering
\small
\caption{\small{Runtime and cost comparison between FEST and Felix on brand text classification (mean $\pm$ std over 5 brands), using GPT-4o-mini \citep{openai2024gpt4omini} as the LLM backbone. FEST is 86$\times$ cheaper, 4.4$\times$ faster, and consumes 91$\times$ fewer tokens. FEST's efficiency stems from two complementary mechanisms: semantic consolidation reduces the feature space early via clustering, and iterative pruning eliminates low-discriminability features so they are never re-encoded in subsequent iterations.}}
\begin{tabular}{lcc}
\toprule
\textbf{Metric} & \textbf{FEST (Ours)} & \textbf{Felix} \\ \midrule
Cost per run (USD) & \textbf{\$0.10 $\pm$ 0.03} & \$8.62 $\pm$ 4.64 \\
Runtime (minutes)  & \textbf{15.9 $\pm$ 3.2}   & 69.4 $\pm$ 35.9 \\
Total tokens       & \textbf{75K $\pm$ 20K}    & 6.8M $\pm$ 3.5M \\ \midrule
Cost ratio  & \multicolumn{2}{c}{\textbf{86$\times$ cheaper}} \\
Speed ratio & \multicolumn{2}{c}{\textbf{4.4$\times$ faster}} \\
Token ratio & \multicolumn{2}{c}{\textbf{91$\times$ fewer}} \\ \bottomrule
\end{tabular}
\end{table}

Felix's one-shot generation produces a large feature set that must be fully encoded for every sample, leading to substantially higher token consumption. FEST's runtime is a one-time offline training cost; at inference, only a single LLM pass against 10--15 retained features is needed.

\section{LLM-as-Judge Expert Coverage}
\label{app:llm_judge}

\subsection{Protocol}
For each brand (Adobe, LG, Porsche), we evaluate the top-20 features from FEST and Felix against all expert-designed brand voice characteristics. GPT-4o serves as the judge, rating each (expert guideline, discovered feature) pair on a 0--10 semantic alignment scale. A guideline is ``covered'' if any discovered feature scores at or above the threshold. This removes dependence on embedding similarity cutoffs and provides interpretable, judgment-based quality scores.

\subsection{Full Sensitivity Table}

\begin{table}[H]
\centering
\caption{\small{LLM-as-judge coverage (\%) of expert brand voice characteristics at varying semantic alignment thresholds. FEST coverage is perfectly stable from threshold 5 through 7 (zero change for all 3 brands), confirming covered guidelines are strong matches (7+/10). Felix collapses at each step: 0\% for Porsche at $\geq$7 vs.\ FEST's 60\%.}}
\small
\begin{tabular}{llcccc}
\toprule
\textbf{Brand} & \textbf{Method} & \textbf{$\geq$5} & \textbf{$\geq$6} & \textbf{$\geq$7} & \textbf{$\geq$8} \\ \midrule
\multirow{2}{*}{Adobe}   & FEST (Ours) & \best{80.0} & \best{80.0} & \best{80.0} & \best{40.0} \\
                         & Felix       & 73.3 & 60.0 & 53.3 & 40.0 \\ \midrule
\multirow{2}{*}{LG}      & FEST (Ours) & 69.2 & \best{69.2} & \best{69.2} & \best{61.5} \\
                         & Felix       & \best{84.6} & 69.2 & 53.8 & 38.5 \\ \midrule
\multirow{2}{*}{Porsche} & FEST (Ours) & \best{60.0} & \best{60.0} & \best{60.0} & \best{40.0} \\
                         & Felix       & 60.0 & 10.0 & \textcolor{red!70!black}{\textbf{0.0}} & \textcolor{red!70!black}{\textbf{0.0}} \\ \bottomrule
\end{tabular}
\end{table}

Notably, at the lenient threshold $\geq$5, Felix actually leads on LG (84.6\% vs.\ 69.2\%), but these are weak matches that collapse at stricter thresholds, while all of FEST's matches persist through $\geq$7.

\section{Why FEST Achieves Higher Coverage}
\label{app:coverage_analysis}

Both FEST and Felix use contrastive samples and the same backbone LLM (GPT-4o-mini), so the coverage gap stems from what happens \textit{after} feature proposals. Three architectural differences drive FEST's superior alignment:

\begin{enumerate}[leftmargin=16pt]
\item \textit{Cluster summarization}: Felix selects the feature closest to each cluster centroid, preserving idiosyncratic LLM phrasing. FEST summarizes clusters into canonical descriptions, producing language that converges toward how experts naturally express concepts.

\item \textit{Iterative pruning}: Felix is one-shot, so all features survive regardless of quality. FEST's tree-guided importance scores identify and prune generic features (e.g., ``uses positive language'') that lack discriminative power. Since expert features are inherently discriminative (experts select what distinguishes classes), iterative selection for discriminative power naturally converges toward expert-aligned features.

\item \textit{Multi-stage language refinement}: FEST edits feature language at discovery, during cluster summarization, and during bank merging across iterations. Each consolidation step distills toward more precise, canonical formulations, while Felix's single-pass centroid-pick cannot improve feature framing based on feedback from data.
\end{enumerate}

\section{Uncovered Guidelines Audit}
\label{app:uncovered}

An audit of all expert guidelines not covered by FEST at threshold $\geq$7 reveals they fall into categories that are structurally unobservable from social media post text:

\begin{itemize}[leftmargin=8pt]
    \item \textbf{Organizational policies}: ``Zero-tolerance for hate speech,'' ``DEI copy-editing policy.'' These are internal editorial standards not manifested in the style of published posts.
    \item \textbf{Abstract brand philosophy}: High-level brand values (e.g., ``building a better working world'') that do not translate to measurable textual style features.
    \item \textbf{Platform meta-guidelines}: ``Design messages to fit the platform,'' ``Adapt content for each channel.'' These require cross-platform context unavailable from individual posts.
\end{itemize}

These guidelines receive LLM-judge scores of 4--6/10 (semantically proximate but not strongly covered), indicating that FEST discovers related features in the same semantic neighborhood but cannot fully capture structurally unmeasurable dimensions. This gap is not specific to FEST; no text-only feature generator (FEST, Felix, or zero-shot LLM) could measure these from post content alone.

\section{Synthetic Off-Brand Robustness}
\label{app:synthetic}

A natural concern is that FEST might learn trivial brand-specific entities or slogans rather than genuine voice/style patterns. We address this through both architectural design and empirical validation.

\textbf{Architectural defense.} Exclusion of trivial brand identifiers is a core design choice. It is explicitly enforced at all three LLM call sites in the FEST pipeline:

\begin{itemize}[leftmargin=12pt, noitemsep, topsep=2pt]
\item \textbf{Stage 1 (Generation):} ``Do NOT mention obvious identifiers like brand names, specific products, hashtags, URLs, logos, or location-specific references.''
\item \textbf{Stage 2 (Cluster summarization):} ``Avoids any references to superficial brand identifiers (names, products, hashtags).''
\item \textbf{Stage 3 (Feature inference):} ``Ignore superficial identifiers: Brand names, Product names, Hashtags, URLs, Logos.''
\end{itemize}

These constraints are enforced consistently at every LLM call, making the system structurally incapable of generating or inference on features like ``mentions Porsche'' or ``includes product URL.''

\textbf{Empirical validation.} We further validate by generating synthetic off-brand content using GPT-4o-mini that matches each brand's topics but uses generic writing style (no brand-specific voice). The negative class is topically identical to the brand's real posts; a shortcut learner relying on trivial brand markers would fail.

\begin{table}[H]
\centering
\caption{\small{FEST accuracy on synthetic off-brand content. Topic-matched but style-generic content is generated per brand using GPT-4o-mini. High accuracy confirms FEST captures voice/style, not brand name shortcuts.}}
\begin{tabular}{lcl}
\toprule
\textbf{Brand} & \textbf{Accuracy (\%)} & \textbf{Top discovered features} \\ \toprule
Adobe & 84.4 & instructional tone, practical guidance \\
LG & 79.8 & vivid imagery, sensory language \\
Porsche & 91.2 & emotional storytelling, sentence length variance \\ \bottomrule
\end{tabular}
\end{table}

FEST maintains 79.8--91.2\% accuracy even when the negative class contains the same topics, brand names, and product keywords. The top features are purely stylistic: ``instructional tone focused on practical applications'' (Adobe SE), ``vivid imagery and sensory language to forge emotional connections'' (LG SE), ``sentence length variance'' (DE). No brand name or product mention features appear in the retained feature bank.

\section{Contamination Discussion}
\label{app:contamination}

Since GPT-4o-mini's training data is not publicly documented, we cannot fully rule out data contamination. However, several observations mitigate this concern:

\begin{enumerate}[leftmargin=12pt]
    \item \textbf{Same-LLM baselines}: Zero-Shot and Few-Shot baselines use the identical LLM on the same content. Any memorization benefit applies equally to all methods, yet FEST consistently outperforms these baselines, indicating methodological gains.
    \item \textbf{Contamination-implausible datasets}: FEST achieves strong results on Dreaddit (Reddit posts about stress) and GPT-generated content detection. Reddit posts are unlikely to be memorized in their task-specific labels, and GPT-generated content detection requires distinguishing model outputs from human writing, not recalling training data.
    \item \textbf{Feature interpretability}: FEST's features are expressed as natural-language descriptions or short executable functions and can be inspected for face validity. The discovered features (e.g., ``uses emotional storytelling,'' ``sentence length variance'') are domain-meaningful, not artifacts of memorization.
\end{enumerate}

We acknowledge this as a limitation inherent to all closed-source LLM evaluations and encourage future work with open-source models where training data provenance is verifiable.

\section{Expert Human Study}
\label{app:human_study}

\subsection{Motivation and Expert Recruitment}
Task accuracy measures discriminative power of features but not whether practitioners find features meaningful in practice. We complement the automatic evaluation with a structured expert evaluation study. One domain practitioner per brand with direct professional experience in brand marketing or content strategy rated the FEST feature bank for their assigned brand. Brand-guideline evaluation requires brand-specific institutional knowledge that cannot be crowd-sourced: a large panel of non-specialist annotators would produce low-signal ratings, while a single qualified practitioner provides authoritative practitioner acceptance. This follows the specialist-rater paradigm used in expert-evaluation studies where the relevant expertise is scarce and non-commoditizable.

\subsection{Protocol}
\label{app:protocol_human_study}
For each brand, FEST produces a final feature bank containing both refined expert features and newly discovered features. They are presented to the expert as a \textit{single blinded pool}: the expert does not know which features are refined from expert guidelines and which are newly discovered by FEST. For each feature, the interface displayed up to 10 content samples receiving the highest attribution score for that feature, providing concrete evidence of how the feature manifests in real brand content. Top 15 features were rated per brand across 2 brands (Zomato images, Adobe images). Expert details are anonymized. Experts rated each feature on three dimensions (1--5 Likert scale):

\begin{itemize}[leftmargin=8pt]
    \item \textbf{Relevance}: Does this feature capture something meaningful and important for the brand?
    \item \textbf{Clarity}: Is the feature description precise and unambiguous?
    \item \textbf{Actionability}: Can a practitioner concretely apply this feature to evaluate new brand content?
\end{itemize}

\subsection{Results}

\begin{table}[H]
\centering
\caption{\small{Domain expert ratings of 15 FEST-discovered features per brand (1--5 Likert scale). Features were presented as a single blinded pool interleaving refined expert features and newly FEST-discovered features. All scores exceed 3.5, confirming features are relevant, clear, and actionable to practitioners.}}
\label{tab:expert_ratings}
\begin{tabular}{lccc}
\toprule
\textbf{Brand} & \textbf{Relevance} & \textbf{Clarity} & \textbf{Actionability} \\ \toprule
Zomato (images) & 4.20 & 4.33 & 4.13 \\
Adobe (images)  & 4.04 & 3.91 & 3.80 \\ \bottomrule
\end{tabular}
\end{table}

All scores exceed 3.5 across both brands and all three dimensions (range 3.80--4.33), confirming that FEST features are relevant, clear, and actionable to domain practitioners. The blinded design ensures that these ratings validate FEST's complete output, including both refined expert features and newly discovered data-specific features, without biasing the expert toward either category.
The two brands span distinct industry sectors (enterprise creative software: Adobe; B2C food delivery: Zomato), providing broader coverage than a within-sector study. 

\subsection{Limitations}
The study does not report inter-rater agreement because brand-evaluation expertise is scarce by design: a large panel of non-specialist annotators would produce low-signal ratings, and recruiting multiple independent practitioners per brand is infeasible within standard research constraints. Our design prioritizes depth over breadth, one highly qualified practitioner per brand which is the appropriate methodology when the task requires domain knowledge that cannot be distributed across a crowd.

\section{Hyperparameters}
\label{app:hyperparameters}

\subsection{Temperature Configuration}

\begin{table}[H]
\centering
\caption{\small{Per-component LLM temperature settings. Higher temperatures encourage diversity for feature generation and exploration stages, while near-zero temperatures ensure deterministic encoding during feature inference.}}
\begin{tabular}{lcl}
\toprule
\textbf{Pipeline Stage} & \textbf{Temperature} & \textbf{Rationale} \\ \toprule
SE feature generation & 0.5 & Encourage diverse hypotheses \\
Feature inference & 0.01 & Near-deterministic encoding \\
Cluster summarization & 0.2 & Balanced precision \\
DE feature ideation & 0.7 & Creative exploration \\
DE code generation & 0.1 & Precise implementation \\ \bottomrule
\end{tabular}
\end{table}

\subsection{Other Hyperparameters}
\begin{itemize}[leftmargin=8pt]
    \item Batch size $K$: 50 pairs per iteration
    \item K-means clusters: $k = 30$ (for SE feature deduplication)
    \item Convergence threshold $\tau_{accuracy}$: 0.95
    \item Importance pruning threshold $\tau_{importance}$: 0.04
    \item Prompt templates per pair ($M$): 3
    \item Typical feature counts: $\sim$300 raw SE candidates per iteration $\rightarrow$ 30 after K-means $\rightarrow$ ~15 after DT pruning across iterations; DE features per iteration: 5--10
    \item Embedding model: Qwen3-Embedding-4B for conditional embeddings
    \item Similarity threshold for semantic clustering of features: 0.8
\end{itemize}

\section{Prompt Templates}
\label{app:prompts}

Below we provide representative prompt templates for each FEST pipeline stage. These templates capture the core structure, input/output format, and key constraints; the full production prompts include additional task-specific instructions and formatting details. Placeholders are shown in \texttt{\{braces\}}.

\begin{tcolorbox}[colback=promptbg, colframe=black!40, fonttitle=\small\bfseries, title=P.1~~SE Feature Proposal, fontupper=\small\ttfamily, boxrule=0.5pt, arc=2pt, left=5pt, right=5pt, top=3pt, bottom=3pt]
\textbf{System:} You are a \{domain\} expert specializing in \{task\_description\}. \\[2pt]
\textbf{Input:} \\
~~Positive sample (class +): \{positive\_sample\} \\
~~Negative sample (class --): \{negative\_sample\} \\[2pt]
\textbf{Task:} Identify \{num\_features\} specific, concrete, and verifiable features (15--20 words each) that explain why the positive sample belongs to class~+. Focus on observable, measurable characteristics. \{task\_constraints\} \\[2pt]
\textbf{Output:} JSON \{``features'': [``feature 1'', ``feature 2'', ...]\}
\end{tcolorbox}

In practice, FEST issues both a positive-discriminator prompt (as above) and a symmetric negative-discriminator prompt (identifying why the negative sample fails to match class~+). Both variants use the same template with reversed sample order.
\vspace{-4pt}

\begin{tcolorbox}[colback=promptbg, colframe=black!40, fonttitle=\small\bfseries, title=P.2~~Cluster Summarization, fontupper=\small\ttfamily, boxrule=0.5pt, arc=2pt, left=5pt, right=5pt, top=3pt, bottom=3pt]
\textbf{System:} You are a \{domain\} consultant specializing in defining guidelines. \\[2pt]
\textbf{Input:} A cluster of related features: \{cluster\_features\} \\[2pt]
\textbf{Task:} Create a single representative feature that captures the core characteristic shared by most features in the cluster. Exclude outliers. Keep concise ($\leq$20 words). \{task\_constraints\} \\[2pt]
\textbf{Output:} Plain text (single feature summary)
\end{tcolorbox}
\vspace{-4pt}

\begin{tcolorbox}[colback=promptbg, colframe=black!40, fonttitle=\small\bfseries, title=P.3~~SE Feature Inference, fontupper=\small\ttfamily, boxrule=0.5pt, arc=2pt, left=5pt, right=5pt, top=3pt, bottom=3pt]
\textbf{System:} You evaluate whether features are present in content samples related to \{task\_description\}. \{task\_constraints\} \\[2pt]
\textbf{Input:} \\
~~Sample: \{sample\} \\
~~Features: \{feature\_list\} \\[2pt]
\textbf{Task:} For each feature, indicate whether the sample exhibits it. \\[2pt]
\textbf{Output:} JSON \{0: ``1'', 1: ``0'', 2: ``1'', ...\}
\end{tcolorbox}

\noindent Token log-probabilities for ``1''/``0'' are extracted and normalized to obtain continuous confidence scores (\S\ref{sec:feature_encoding}).
\vspace{-4pt}

\begin{tcolorbox}[colback=promptbg, colframe=black!40, fonttitle=\small\bfseries, title=P.4~~DE Feature Ideation, fontupper=\small\ttfamily, boxrule=0.5pt, arc=2pt, left=5pt, right=5pt, top=3pt, bottom=3pt]
\textbf{System:} You are a data scientist specializing in feature engineering for \{task\_description\}. Propose features that are: (1) objectively computable in Python, (2) deterministic, (3) CPU-efficient. \\[2pt]
\textbf{Input:} \\
~~Positive samples: \{positive\_samples\} \\
~~Negative samples: \{negative\_samples\} \\
~~Existing features (do NOT duplicate): \{existing\_features\} \\[2pt]
\textbf{Task:} Propose up to \{max\_features\} new computable features. For each: name (snake\_case), description, data\_type (numeric$|$boolean), rationale. \\[2pt]
\textbf{Output:} JSON \{``features'': [\{``name'': ..., ``description'': ..., ``data\_type'': ..., ``rationale'': ...\}, ...]\}
\end{tcolorbox}
\vspace{-4pt}

\begin{tcolorbox}[colback=promptbg, colframe=black!40, fonttitle=\small\bfseries, title=P.5~~DE Code Generation, fontupper=\small\ttfamily, boxrule=0.5pt, arc=2pt, left=5pt, right=5pt, top=3pt, bottom=3pt]
\textbf{System:} You are an expert Python developer specializing in feature extraction. \\[2pt]
\textbf{Input:} Feature name: \{name\}, Description: \{description\}, Data type: \{data\_type\}, Input modality: \{input\_modality\} \\[2pt]
\textbf{Task:} Write a Python function \texttt{extract\_feature(\{signature\})} that returns a single value. Handle edge cases gracefully. Allowed libraries: \{allowed\_imports\}. No ML models or external APIs. \\[2pt]
\textbf{Output:} Python code only (no markdown fences).
\end{tcolorbox}

\noindent Generated code is sandbox-validated (compiled, executed on sample data, output-type checked) before inclusion in the feature bank.

\section{BrandGuide Dataset Details}
\label{app:dataset}

This appendix provides comprehensive details on the BrandGuide dataset, including our collection methodology, quality assurance procedures, extended statistics, and representative examples.

\subsection{Collection Pipeline}

Our multi-stage pipeline combines automated extraction with rigorous quality control to ensure dataset integrity:
\begin{enumerate}[leftmargin=*,noitemsep,topsep=0pt]
    \item \textbf{Data Acquisition}: We systematically collected brand guidelines from the web, extracting rich metadata including publication year, geographic region, language, and sector tags. Initial collection yielded 3,466 candidate entries spanning 1963--2025.
    
    \item \textbf{Temporal Filtering}: To ensure contemporary relevance and consistency in design conventions, we filtered entries to the 2014--2025 timeframe, yielding 2,683 brands that reflect modern digital-first brand systems.
    
    \item \textbf{Guideline Extraction}: Each document undergoes structured parsing to extract design specifications as text including color codes (HEX, RGB, CMYK, Pantone), typography hierarchies (primary/secondary typefaces, weights, sizes), logo clearance rules (minimum sizes, spacing requirements), and usage constraints (approved/prohibited applications).
    
    \item \textbf{Visual Asset Retrieval}: For each brand, we retrieve imagery through web search using brand name and relevant keywords. We collect real-world logo applications, color implementations, marketing collateral, and brand touchpoints. This process yielded approximately \textbf{1M} brand images and textual descriptions across all entries.
    
    \item \textbf{Manual Verification}: Each stage incorporates human review to ensure annotation accuracy, filter malformed entries, and validate guideline-image alignment. Annotators verified that extracted specifications match source documents and that retrieved images accurately represent the corresponding brand.
\end{enumerate}

\subsection{Quality Assurance}

To maintain dataset quality, we implemented several verification procedures:
\vspace{-0.25em}
\begin{itemize}[leftmargin=*,noitemsep,topsep=0pt]
    \item \textbf{Specification Validation}: Extracted color codes were validated against standard formats; typography specifications were checked for completeness.
    \item \textbf{Image Filtering}: Retrieved images underwent automated filtering for resolution, followed by manual review for ambiguous cases.
    \item \textbf{Duplicate Detection}: We removed duplicate brands and near-duplicate guideline versions, retaining the most recent edition for each brand.
    \item \textbf{Metadata Verification}: Publication years, regions, and sector tags were cross-referenced with source documents and corrected where inconsistencies were detected.
\end{itemize}

\subsection{Extended Statistics}

\noindent\textbf{Geographic Distribution.} The dataset exhibits strong international coverage with representation from 103 regions. While USA (35.0\%), UK (9.8\%), and France (6.7\%) comprise the largest segments, substantial coverage spans Europe (Germany, Spain, Italy, Netherlands, Switzerland), Asia (Japan, India, Indonesia, China), and Latin America (Brazil, Colombia, Mexico). This diversity enables cross-cultural analysis of design conventions and regional branding patterns.

\noindent\textbf{Sector Diversity.} Guidelines span 80 sectors including Education (385), Sport (230), Technology (144), Software (134), Food \& Beverage (117), and Financial Services (104). The sector distribution reflects real-world brand guideline availability, with educational institutions and sports organizations particularly well-represented due to their public communication requirements. This diversity enables domain-specific analysis of design conventions across organizational types.

\noindent\textbf{Temporal Coverage.} With guidelines spanning 2014--2025, BrandGuide captures contemporary design trends during the era of digital-first brand systems, responsive identity design, and the rise of design systems. The distribution peaks in 2019 (375 brands) and shows consistent coverage across years, enabling longitudinal studies of evolving design practices.

\noindent\textbf{Language Distribution.} English dominates (79.4\%), reflecting global business practices, but the dataset includes substantial multilingual coverage: French (153), Spanish (108), Portuguese (47), German (37), and Arabic (27), among 28 total languages. This enables research on language-specific design conventions and cross-lingual brand communication.

\begin{table}[h]
\centering
\caption{BrandGuide dataset overview: 2,683 brand guidelines across 80 sectors, 103 regions, and 28 languages (2014--2025).}
\label{tab:dataset_full}
\small

\vspace{0.5em}
\textbf{(a) Summary Statistics}
\vspace{0.3em}

\begin{tabular}{@{}lrlrlr@{}}
\toprule
\textbf{Statistic} & \textbf{Value} & \textbf{Statistic} & \textbf{Value} & \textbf{Statistic} & \textbf{Value} \\
\midrule
Total brands & 2,683 & Geographic regions & 103 & Year span & 2014--2025 \\
Sectors & 80 & Languages & 28 & Total images & $\sim$1M \\
\bottomrule
\end{tabular}

\vspace{1em}
\textbf{(b) Distribution by Sector, Region, and Language}
\vspace{0.3em}

\begin{tabular}{@{}lrlrlr@{}}
\toprule
\textbf{Sector} & \textbf{\#} & \textbf{Region} & \textbf{\#} & \textbf{Language} & \textbf{\#} \\
\midrule
Education & 385 & USA & 940 & English & 2,129 \\
Sport & 230 & United Kingdom & 262 & French & 153 \\
Regional & 202 & France & 179 & Spanish & 108 \\
Corporate & 186 & International & 134 & Portuguese & 47 \\
Technology & 144 & Canada & 126 & German & 37 \\
Software & 134 & Australia & 73 & Arabic & 27 \\
Food \& Beverage & 117 & Spain & 67 & Italian & 26 \\
Transport & 109 & Germany & 64 & Russian & 20 \\
Events & 106 & Italy & 45 & Chinese & 20 \\
Financial & 104 & India & 40 & Japanese & 16 \\
NGO & 96 & Japan & 39 & Indonesian & 15 \\
Tourism & 87 & Ireland & 37 & Catalan & 13 \\
\midrule
Others (68) & 783 & Others (91) & 677 & Others (16) & 72 \\
\bottomrule
\end{tabular}

\vspace{1em}
\textbf{(c) Temporal Distribution}
\vspace{0.3em}

\begin{tabular}{@{}cccccccccccc@{}}
\toprule
\textbf{2014} & \textbf{2015} & \textbf{2016} & \textbf{2017} & \textbf{2018} & \textbf{2019} & \textbf{2020} & \textbf{2021} & \textbf{2022} & \textbf{2023} & \textbf{2024} & \textbf{2025} \\
\midrule
144 & 203 & 251 & 311 & 321 & 375 & 295 & 267 & 173 & 148 & 133 & 62 \\
\bottomrule
\end{tabular}
\end{table}

\subsection{Research Directions}
\textbf{BrandGuide} supports multiple research directions: (1)~\textit{brand consistency verification}: automated compliance checking of design assets against guidelines; (2)~\textit{generative design}: training models to produce brand-coherent visual assets conditioned on textual specifications; (3)~\textit{design trend analysis}: studying temporal and geographic patterns in visual identity; and (4)~\textit{multimodal grounding}: learning alignments between natural language design descriptions and precise visual properties. 

\subsection{Dataset Examples}

We provide representative examples demonstrating the structure and content of BrandGuide entries. Each example illustrates how expert-authored design specifications are paired with corresponding visual assets.

\noindent\textbf{Example Structure.} Each brand entry contains:
\begin{itemize}[leftmargin=*,noitemsep,topsep=0pt]
    \item \textit{Brand metadata}: name, sector, region, language, publication year
    \item \textit{Visual assets}: logo files, brand imagery
\end{itemize}

\noindent\textbf{Representative Samples.} We include two complete brand examples in the supplementary materials:
\begin{itemize}[leftmargin=*,noitemsep,topsep=0pt]
    \item \texttt{LG}: A global technology and electronics brand demonstrating comprehensive digital-first guidelines with detailed color systems, responsive logo variants, and extensive application rules across product categories.
    \item \texttt{Porsche}: A luxury automotive brand showcasing premium brand architecture with precise typography hierarchies, strict color specifications, and meticulous guidelines for maintaining brand prestige across touchpoints.
\end{itemize}

\noindent These examples illustrate the diversity of design approaches across sectors and the granularity of expert specifications captured in BrandGuide. Each folder contains the extracted guideline text, and corresponding visual assets.

\subsection{Licensing and Access}

BrandGuide will be released for non-commercial research purposes. The dataset provides structured access to brand guidelines and associated imagery; all underlying intellectual property rights remain with the respective brand owners and guideline authors. We do not claim ownership over any third-party brand assets included in the dataset.
To ensure transparency and provenance, we will release: (i) a compiled attribution list identifying the authors of all brand guidelines, (ii) image URLs rather than raw image files, enabling independent provenance verification and allowing rights holders to request removal if needed.
Users of BrandGuide are required to restrict usage to non-commercial academic research. Any other intended use must be communicated to the authors prior to deployment. By accessing the dataset, users acknowledge that compliance with the terms of the original copyright holders remains their responsibility.

\section{LLM Usage}
Large Language Models were used solely as a writing assistance tool during the preparation of this manuscript. Specifically, LLMs were employed to: ~(1) Polish and refine the language and clarity of written sections ~(2) Assist with formatting and organization of content. LLMs were \textit{not} involved in any aspect of the research ideation, methodology design, experimental design, data analysis or interpretation of results.

\section{Feature Examples}

\begin{longtable}{|p{0.44\linewidth}|p{1.8cm}|p{0.30\linewidth}|}
\caption{\small{Representative features discovered by FEST from brand social media post \textbf{text} (subset of the full feature bank, selected for qualitative inspection). SE = Semantic (LLM-scored); DE = Deterministic (executable Python function).}}\label{tab:qualitative_text}\\
\hline
\rowcolor{blue!8}\cellcolor{blue!8}\textbf{Feature} & \textbf{Type} & \textbf{Code} \\ \hline
\endfirsthead
\hline
\rowcolor{blue!8}\cellcolor{blue!8}\textbf{Feature} & \textbf{Type} & \textbf{Code} \\ \hline
\endhead

\multicolumn{3}{|l|}{\cellcolor{gray!18}\textbf{Louis Vuitton}} \\ \hline
Authoritative, sophisticated tone celebrating cultural figures in high-fashion discourse. & SE & --- \\ \cline{1-3}
Formal vocabulary and complex sentence structures conveying luxury and exclusivity. & SE & --- \\ \cline{1-3}
Average sentence length (words per sentence). & DE & \begin{lstlisting}[language=Python]
def extract_feature(text: str):
    if not text or not isinstance(text, str):
        return 0.0
    sentences = nltk.sent_tokenize(text)
    if not sentences:
        return 0.0
    total_words = sum(
        len(nltk.word_tokenize(s))
        for s in sentences)
    return total_words / len(sentences)
\end{lstlisting} \\ \cline{1-3}
Average number of clauses per sentence (comma/semicolon delimited). & DE & \begin{lstlisting}[language=Python]
def extract_feature(text: str):
    if not text or not isinstance(text, str):
        return 0.0
    sentences = nltk.sent_tokenize(text)
    if not sentences:
        return 0.0
    clause_count = sum(
        len([c for c in re.split(r'[;,]', s)
             if c.strip()])
        for s in sentences)
    return clause_count / len(sentences)
\end{lstlisting} \\ \hline

\multicolumn{3}{|l|}{\cellcolor{gray!18}\textbf{Emirates}} \\ \hline
Clear, friendly short-form language for broad audience connection. & SE & --- \\ \cline{1-3}
Informative value propositions emphasizing travel experiences and service quality. & SE & --- \\ \cline{1-3}
Frequency of @-mentions of travel partners and destination accounts. & DE & \begin{lstlisting}[language=Python]
def extract_feature(text: str):
    if not isinstance(text, str) or text is None:
        return 0
    mentions = re.findall(r'@\w+', text)
    return len(mentions)
\end{lstlisting} \\ \cline{1-3}
Average character length of hashtags used in the post. & DE & \begin{lstlisting}[language=Python]
def extract_feature(text: str):
    if not text or not isinstance(text, str):
        return 0.0
    hashtags = re.findall(r'#\w+', text)
    if not hashtags:
        return 0.0
    return (sum(len(h) for h in hashtags)
            / len(hashtags))
\end{lstlisting} \\ \hline

\multicolumn{3}{|l|}{\cellcolor{gray!18}\textbf{Pizza Hut}} \\ \hline
Urgency-driven language with compelling calls-to-action for limited-time offers. & SE & --- \\ \cline{1-3}
Vivid sensory language and emotional storytelling evoking food cravings. & SE & --- \\ \cline{1-3}
Ratio of emoji characters to total post length. & DE & \begin{lstlisting}[language=Python]
def extract_feature(text: str):
    if text is None or not isinstance(text, str):
        return 0.0
    total_chars = len(text)
    if total_chars == 0:
        return 0.0
    emoji_count = sum(
        1 for c in text
        if unicodedata.category(c).startswith('So'))
    return emoji_count / total_chars
\end{lstlisting} \\ \cline{1-3}
Ratio of exclamation marks to total punctuation marks. & DE & \begin{lstlisting}[language=Python]
def extract_feature(text: str):
    if not text or not isinstance(text, str):
        return 0.0
    punc = sum(1 for c in text
               if c in string.punctuation)
    if punc == 0:
        return 0.0
    return text.count('!') / punc
\end{lstlisting} \\ \hline

\end{longtable}

\begin{longtable}{|p{0.44\linewidth}|p{1.8cm}|p{0.30\linewidth}|}
\caption{\small{Representative features discovered by FEST from brand promotional \textbf{images} (subset of the full feature bank, selected for qualitative inspection). SE = Semantic (LLM-scored); DE = Deterministic (executable Python function).}}\label{tab:qualitative_image}\\
\hline
\rowcolor{blue!8}\cellcolor{blue!8}\textbf{Feature} & \textbf{Type} & \textbf{Code} \\ \hline
\endfirsthead
\hline
\rowcolor{blue!8}\cellcolor{blue!8}\textbf{Feature} & \textbf{Type} & \textbf{Code} \\ \hline
\endhead

\multicolumn{3}{|l|}{\cellcolor{gray!18}\textbf{Porsche}} \\ \hline
Sleek dynamic shapes and bold colors showcasing automotive craftsmanship and precision. & SE & --- \\ \cline{1-3}
Dynamic angles and action-oriented compositions conveying speed and performance. & SE & --- \\ \cline{1-3}
Aspect ratio (width/height) capturing widescreen landscape framing for car photography. & DE & \begin{lstlisting}[language=Python]
def extract_feature(image_path: str):
    if not image_path or not isinstance(image_path, str):
        return None
    try:
        image = PIL.Image.open(image_path)
        width, height = image.size
        if height == 0:
            return None
        return width / height
    except (FileNotFoundError, OSError):
        return None
\end{lstlisting} \\ \cline{1-3}
Density of edges detected via Canny operator (sharp, precise automotive lines). & DE & \begin{lstlisting}[language=Python]
def extract_feature(image_path: str):
    if not image_path or not isinstance(image_path, str):
        return 0.0
    if not os.path.isfile(image_path):
        return 0.0
    try:
        img = cv2.imread(image_path,
                         cv2.IMREAD_GRAYSCALE)
        if img is None:
            return 0.0
        edges = cv2.Canny(img, 100, 200)
        return cv2.countNonZero(edges) / img.size
    except Exception:
        return 0.0
\end{lstlisting} \\ \hline

\multicolumn{3}{|l|}{\cellcolor{gray!18}\textbf{Pizza Hut}} \\ \hline
Vibrant warm color palette (red-dominant) evoking appetite and energy. & SE & --- \\ \cline{1-3}
Bold playful typography with festive visual elements and vibrant colors. & SE & --- \\ \cline{1-3}
Luminance-weighted color contrast ratio between dominant and background color. & DE & \begin{lstlisting}[language=Python]
def extract_feature(image_path: str):
    if not image_path or not isinstance(image_path, str):
        return 0.0
    try:
        img = PIL.Image.open(image_path).convert("RGB")
    except Exception:
        return 0.0
    pixels = list(img.getdata())
    if not pixels:
        return 0.0
    dominant = collections.Counter(pixels).most_common(1)[0][0]
    background = pixels[0]
    def lum(c):
        return 0.2126*c[0] + 0.7152*c[1] + 0.0722*c[2]
    L1, L2 = lum(dominant), lum(background)
    lo, hi = min(L1, L2), max(L1, L2)
    return (hi + 0.05) / (lo + 0.05)
\end{lstlisting} \\ \cline{1-3}
Count of distinct visual elements via HSV color segmentation (food variety and abundance). & DE & \begin{lstlisting}[language=Python]
def extract_feature(image_path: str):
    if not image_path or not isinstance(image_path, str):
        return 0
    if not os.path.isfile(image_path):
        return 0
    try:
        image = cv2.imread(image_path)
        if image is None:
            return 0
        hsv = cv2.cvtColor(image, cv2.COLOR_BGR2HSV)
        lo = np.array([0, 50, 50])
        hi = np.array([180, 255, 255])
        mask = cv2.inRange(hsv, lo, hi)
        contours, _ = cv2.findContours(
            mask, cv2.RETR_EXTERNAL,
            cv2.CHAIN_APPROX_SIMPLE)
        return len(contours)
    except Exception:
        return 0
\end{lstlisting} \\ \hline

\multicolumn{3}{|l|}{\cellcolor{gray!18}\textbf{Louis Vuitton}} \\ \hline
Warm sophisticated palette (browns, golds, creams) evoking timeless luxury and elegance. & SE & --- \\ \cline{1-3}
High-quality artistic fashion photography with elegant, sophisticated compositions. & SE & --- \\ \cline{1-3}
Horizontal symmetry score via grayscale histogram intersection of left/right halves. & DE & \begin{lstlisting}[language=Python]
def extract_feature(image_path: str):
    if not image_path or not isinstance(image_path, str):
        return 0.0
    try:
        image = PIL.Image.open(image_path).convert('L')
    except (FileNotFoundError, IOError):
        return 0.0
    width, height = image.size
    if width == 0 or height == 0:
        return 0.0
    left = image.crop((0, 0, width // 2, height))
    right = image.crop(
        (width // 2, 0, width, height)
    ).transpose(PIL.Image.FLIP_LEFT_RIGHT)
    lh = collections.Counter(left.getdata())
    rh = collections.Counter(right.getdata())
    score = sum(min(lh.get(p, 0), rh.get(p, 0))
                for p in set(lh) | set(rh))
    max_score = left.size[0] * left.size[1]
    return score / max_score if max_score > 0 else 0.0
\end{lstlisting} \\ \cline{1-3}
Euclidean distance of bright-pixel centroid from image center (product subject offset). & DE & \begin{lstlisting}[language=Python]
def extract_feature(image_path: str):
    if not image_path or not isinstance(image_path, str):
        return 0.0
    if not os.path.isfile(image_path):
        return 0.0
    try:
        img = PIL.Image.open(image_path).convert("L")
        w, h = img.size
        px = np.array(img)
        fg = px > np.mean(px)
        ys, xs = np.where(fg)
        if len(xs) == 0:
            return 0.0
        cx, cy = np.mean(xs), np.mean(ys)
        return math.sqrt((cx - w/2)**2 + (cy - h/2)**2)
    except Exception:
        return 0.0
\end{lstlisting} \\ \hline

\end{longtable}

\section{Licensing for Existing Assets}
\label{app:assets_license}

\begin{itemize}[leftmargin=16pt]
\item \textbf{GPT-GC} \citep{zhou2024hypothesis}, used for content authenticity detection: MIT License.
\item \textbf{Dreaddit} \citep{turcan2019dreaddit}, used for stress detection: distributed by the authors for research purposes via the ACL Anthology (DOI: 10.18653/v1/D19-6213) and their institutional page at Columbia University. No explicit open-source license is assigned; we use it solely for non-commercial academic research, consistent with its intended distribution and community norms.
\item \textbf{Engaging ImageNet} \citep{khurana2025measuring}: CC BY-NC-ND 4.0.
\end{itemize}

\end{document}